\documentclass{article}

\usepackage{microtype}
\usepackage{graphicx}
\usepackage{booktabs} 

\usepackage{hyperref}

\usepackage[accepted]{mlsys2026}

\usepackage[utf8]{inputenc} 
\usepackage[T1]{fontenc}    
\usepackage{url}            
\usepackage{nicefrac}       
\usepackage{microtype}      
\usepackage{soul}
\usepackage{graphicx}
\usepackage{outlines}
\usepackage{xurl}

\usepackage{amsmath,amsfonts,amsbsy,latexsym}
\usepackage{derivative}
\usepackage[normalem]{ulem}
\usepackage{mathtools}

\usepackage{booktabs}       
\usepackage{multirow}
\usepackage{paralist}
\usepackage{multicol}
\usepackage{diagbox}
\usepackage{makecell}
\usepackage{xcolor}
\usepackage{colortbl}
\usepackage{tabularx}
\usepackage{longtable}

\usepackage{subfig}

\usepackage{algorithm}
\renewcommand{\algorithmicrequire}{\textbf{Input:}}
\renewcommand{\algorithmicensure}{\textbf{Output:}}
\renewcommand{\algorithmiccomment}[1]{\hfill$\triangleright$ \textit{#1}}
\usepackage{listings}
\setboolean{ALC@noend}{true}

\usepackage{enumitem}
\usepackage{amssymb}

\usepackage{url}
\usepackage{color}

\newcommand{\cb}{} 

\newcommand{\cgray}{\textcolor{gray}}

\def \SYS{BatchLLM}

\mlsystitlerunning{\SYS{}: Optimizing Large Batched LLM Inference with Global Prefix Sharing and Throughput-oriented Token Batching}

\begin{document}


\twocolumn[
\mlsystitle{\SYS{}: Optimizing Large Batched LLM Inference with Global Prefix Sharing and Throughput-oriented Token Batching}



\mlsyssetsymbol{equal}{}

\begin{mlsysauthorlist}
\mlsysauthor{Zhen Zheng}{msft}
\mlsysauthor{Xin Ji}{msft}
\mlsysauthor{Taosong Fang}{msft,ucas}
\mlsysauthor{Fanghao Zhou}{msft}
\mlsysauthor{Chuanjie Liu}{msft}
\mlsysauthor{Gang Peng}{msft}
\end{mlsysauthorlist}

\mlsysaffiliation{msft}{Microsoft}
\mlsysaffiliation{ucas}{Chinese Information Processing Laboratory, ISCAS, UCAS. Work was done when Taosong Fang was interned at Microsoft}

\mlsyscorrespondingauthor{Zhen Zheng}{zhengzhen@microsoft.com}


\vskip 0.3in

\begin{abstract}

Large language models (LLMs) increasingly play an important role in a wide range of information processing and management tasks in industry.
Many of these tasks are performed in large batches or even offline, and the performance \cb{indicator} for which is throughput.
These tasks usually show the characteristic of prefix sharing, where different prompt input can partially show the common prefix.
However, the existing LLM inference engines tend to optimize the streaming requests and show limitations of supporting the large batched tasks with the prefix sharing characteristic.
The existing solutions use the LRU-based cache to reuse the KV context of common prefix between requests.
The KV context that are about to be reused \cb{may be prematurely evicted} with the implicit cache management.
Besides, the streaming oriented systems do not leverage the request-batch information and can not mix the decoding tokens with the prefill chunks to the best for the batched scenarios, and thus fails to saturate the GPU.
We propose \SYS{} to address the above problems.
\SYS{} explicitly identifies the common prefixes globally.
The requests sharing the same prefix will be scheduled together to reuse the KV context the best.
\SYS{} reorders the requests and schedules the requests with larger ratio of decoding first to better mix the decoding tokens with the latter prefill chunks, and applies memory-centric token batching to enlarge the token-batch sizes,
which helps to increase the GPU utilization.
Extensive evaluation shows that \SYS{} outperforms vLLM and SGLang by $1.3\times$ to $10.8\times$ on a set of microbenchmarks and a typical industry workload under different hardware environments.
\cb{
Code is available at \url{https://github.com/microsoft/MixLLM/tree/batchllm_vllm_064}.
}

\end{abstract}
]



\printAffiliationsAndNotice{}  


\section{Introduction}


The modern information processing and management tasks are starting to leverage large language models (LLMs) to achieve better results, such as the search engine~\cite{large-search-model, search-engine-meet-llm}, advertisement~\cite{meguellati2024how}, and recommender systems~\cite{zhao2024recommender, kim2024large}.
Due to the huge amount of data to be processed, many of these LLM tasks are performed in large batches or even offline (e.g., offline ranking), for which the most important performance indicator is throughput.
Besides, different input (prompts) of these tasks can partially share the same prefix (e.g., the same web document in the search engine tasks).
Given the high cost of LLM inference, how to serve these tasks efficiently is a critical problem for the information processing systems.


The recent LLM inference engines~\cite{vllm,sglang} can flexibly support the execution of LLM inference tasks, especially with blocked KV memory management~\cite{vllm} and per-iteration token batching~\cite{orca}.
However, the design of existing inference engines tend to optimize streaming online services and show limitations in throughput-oriented large-batch processing.
1) \textit{Lack of global prefix sharing.}
The state-of-the-art (SOTA) LLM inference engines support to cache the KV context with LRU policy to enable prefix sharing between prompts.
However, from a global perspective of large batches, an LRU-based cache may prematurely evict KV contexts that are about to be reused and retain unused KV contexts for a long time, causing unnecessary recalculation.
2) \textit{Suboptimal token batching for throughput-oriented and prefix-shared tasks.}
The current LLM inference engines schedule different requests independently and do not cluster the requests with common prefix together to schedule.
It can extend the lifetime of the common prefix and thus increase the memory usage, and may prematurely evict KV contexts that are about to be reused as discussed above.
Besides, they usually schedule the requests in the first-come-first-serve or similar order to guarantee the fairness and latency of the streaming requests.
The requests with larger ratio of decoding steps may be scheduled too late to be able to mix with the prefill chunks to increase the hardware utilization.
They apply the mix of prefill and decoding tokens together according to the threshold of token number and request number.
This limits the number of decoding tokens that can be batched and keeps the GPUs from saturating for the iterations dominated by decoding tokens.




In this work, we introduce \SYS{}, the holistic optimization for the throughput-oriented large-batch LLM inference with global prefix preprocessing, and prefix-aware and throughput-oriented token batching.
The basic insight is based on the fact that, the prompt characteristics are known ahead before processing the large batch.
Compared to the implicit LRU cache that cannot retain the KV context for reuse accurately, the common prefixes between prompts can be identified explicitly and can be reused explicitly.
Besides, the length distribution of the prompts is known ahead and the main performance target is throughput, which allows the inference engine to reorder the requests and form larger token-batches\footnote{
This paper uses the term \textit{token-batch} to represent a batch of tokens to be processed for an iteration of continuous batching~\cite{orca}. It differs from the concept of \textit{large batch} of all the prompt to be processed in a job.
}.
It makes the following contributions:

$\blacktriangleright$ It pioneers the LLM inference optimization for the throughput oriented large batched prefix-shared scenarios with the insight of using global information of the batch, 
for the increasingly important LLM-based information processing and management.

$\blacktriangleright$ It designs the ahead-of-time prefix identification and enlargement method to achieve effective prefix reusing in global view, 
and proposes the requests reordering and memory-centric token batching method to increase the per-iteration GPU utilization. 

$\blacktriangleright$ It presents \SYS{} implementation building upon vLLM. 
\SYS{} achieves end-to-end performance improvement of $1.3\times$ to $10.8\times$ than vLLM and SGLang for the benchmarks and an industry workload under different hardware environments.


\section{Opportunities}
\label{sec:opportunities}

\subsection{Emerging Industry Scenarios}
\label{subsec:scenarios}

LLM has been increasingly used in a broad range of industry tasks with its in-context learning and reasoning ability~\cite{large-search-model, search-engine-meet-llm, meguellati2024how, zhao2024recommender, kim2024large, lingua-manga, how-llm}.
Due to the large traffic of these tasks and the high cost of LLM inference, many of these tasks are processed in very large batch (or even offline) whose performance metric is mainly the throughput rather than latency.
Take the snippet generation task~\cite{large-search-model, search-engine-meet-llm} in search engine as an example, which is to generate the snippet of the web document (web content) in the search result page according to both document and user query.
An example prompt in industry~\cite{large-search-model} is: \textit{generate a short snippet based on the given Document to answer the given Query. Document: <...> Query: <...>}.
Due to that it is hard to meet the SLO when performing the massive LLM inferences online, a common practice is to run the snippet generation task offline for the high-frequency web documents and queries, and retrieve the offline results during online serving when the corresponding document-query pair occurs.

Many of the LLM-based information processing and management tasks show the characteristic of the shared prefix between the prompts of different requests,
which comes from the nature that different tasks can perform on the same content (e.g., web document).
Take the search engine task as an example, it may extract different information from the same web document so that the document can be a shared prefix in the prompt.
Specifically, as for the snippet generation task, given that it is to extract the snippet of the document-query pair, a document can be the common prefix for different queries.
There are many other industry tasks that fall into the offline processing mode with a common prefix.
The very important offline ranking tasks for search and recommendation can use the prompt of \textit{"please evaluate the relevance of the document for the query, Query: <...>, Document: <...>"}, where each query corresponds to many documents and can be the common prefix.
The Ads title rewrite task in industry can use the prompt of \textit{"given the Query and the Landing Page info, please refine the old title ‘xxx’ to get a better ad title. Landing Page: <...>, Query: <...>"}, where the landing page can be the common prefix and the computation is usually throughput critical.
The offline labeling and scoring tasks in industry can have long system prompt and few shot examples, which can also be the common prefix in the prompt.

The shared prefix can be managed in multiple levels.
For example, the first level prefix can be the global system information and instructions, and the second level can be the web document.
However, with the development of LLM and the using of task-specific model fine-tuning, the prefix of instructions becomes shorter and shorter:
On the one hand, the LLM models are incorporating more and more reasoning abilities into themselves (like OpenAI o1~\cite{openai-o1}) and will not rely on complex prompts.
On the other hand, the task-specific fine-tuning can help to incorporate the complex prompts into the model weight.
As a result, the prefix sharing of the long context rather than the instructions can be the most important for many industry tasks.
This motivates the first-level prefix enlargement in Sec.\ref{subsec:prefix-identification}.

\subsection{The New Optimization Demands}
\label{subsec:limitations}

Even though prefix sharing and per-iteration token batching are functionally supported in previous works, they lack specialized optimization for the increasingly import large batched scenarios (Sec.\ref{subsec:scenarios}),
showing two main limitations:

\textbf{1. The implicit prefix caching cannot achieve the optimal KV reuse for the large batched inference.}
The industrial LLM serving systems~\cite{vllm, sglang} use the prefix tree to maintain the KV blocks.
The same prefix (identified by runtime hashing) can be mapped to the same KV blocks to avoid repeated computation.
It uses the LRU policy to evict the blocks from the block table.
As for a large batch of inputs to be processed, it does not keep the prefix in memory in the global view in the large batch, but only according to the history input.
As a result, the shareable KV blocks can be easily evicted prematurely.
We have evaluated vLLM's prefix sharing with a typical industry workload.
The optimal token saving ratio by prefix reusing is 58.1\% by analyzing the input dataset.
Note that the saving ratio is calculated by:
\begin{equation}
R_{saving} = (1 - \frac{N_{processed\_pre\_tok}}{N_{logical\_pre\_tok}}) \times 100
\label{eq:saving-ratio}
\end{equation}
$N_{processed\_pre\_tok}$ is the number of prefill tokens processed after prefix sharing, and $N_{logical\_pre\_tok}$ is the original prefill token number of all the requests.
The vLLM's implicit prefix caching only achieves 35.8\% (Sec.\ref{subsubsec:breakdown-prefix-grouping}).
There is a need for better prefix sharing for the large batched tasks.

\begin{figure}[]
    \centering
    \includegraphics[width=0.7\columnwidth]{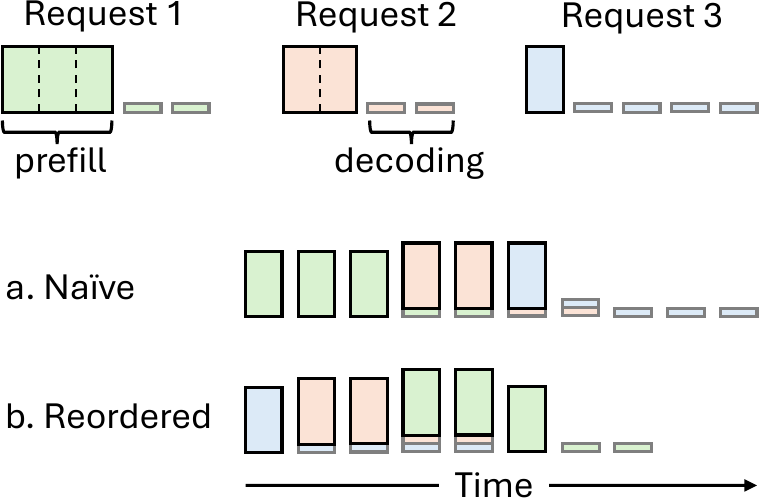}
    \caption{The effect of processing order of requests with chunked-prefill enabled. Given the three requests with different prefill/decoding length characteristics, the naive token batching in the coming order of the requests has worse token mixing of decoding and prefill chunks.}
    \label{fig:processing-order}
\end{figure}

\begin{figure}[]
    \centering
    \includegraphics[width=.95\columnwidth]{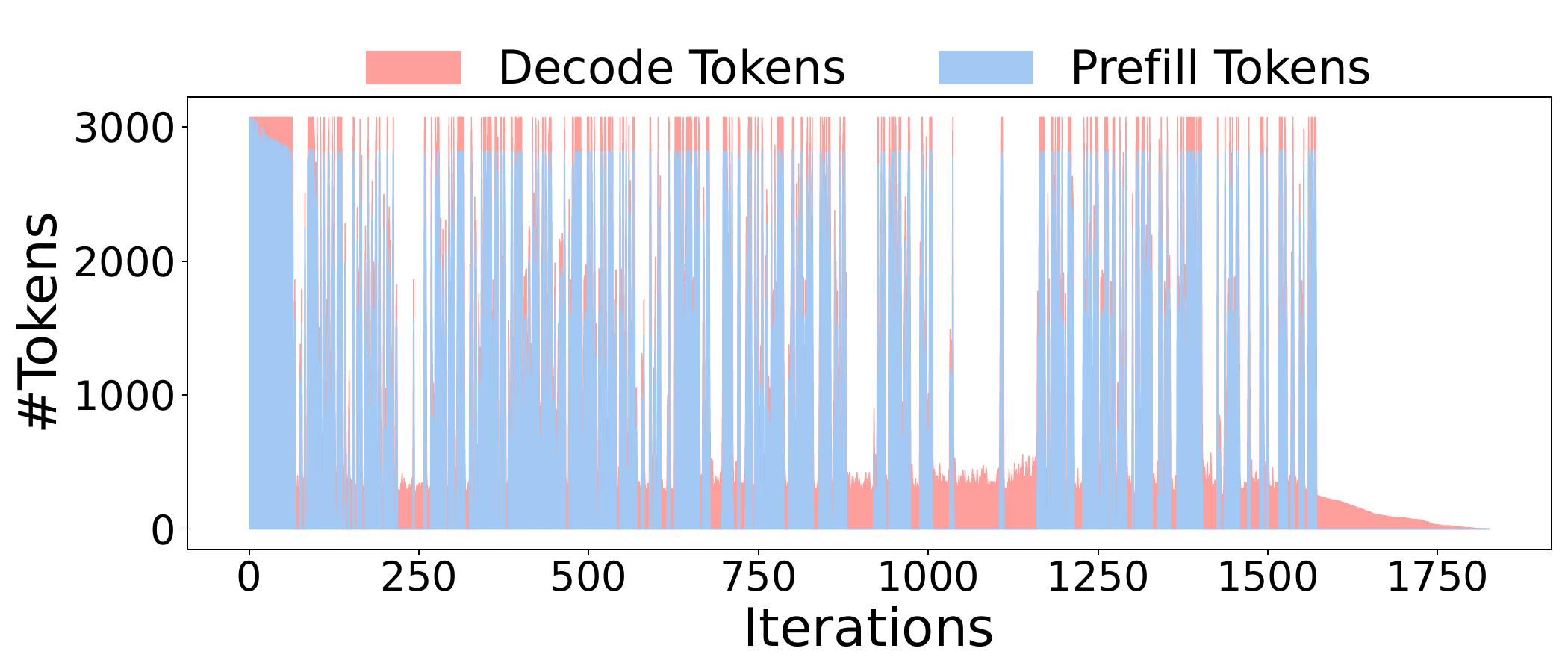}
    \caption{The token number in the batch processed at each iteration for an industry task with vLLM's chunked-prefill. It has "valleys" for many iterations.
    }
    \label{fig:valley-example}
\end{figure}

\textbf{2. The online serving oriented scheduling and token batching can be suboptimal for the prefix-shared and throughput oriented tasks.}
The current LLM inference systems schedule the tasks at the granularity of request.
They do not analyze the prefix sharing characteristics of the whole large batch nor schedule the requests sharing common prefix together.
This not only can evict the shareable KV context prematurely as discussed above, but also extends the lifetime of the shareable KV memory, exacerbating the problem of large KV memory consumption.
Besides, these systems design tends to support online services.
They form the token-batches at each iteration with the constraint of requests' arriving order, which can result in suboptimal token-batches.
Take the example in Fig.\ref{fig:processing-order}, the naive scheduling in the order of request's coming cannot mix the decoding of \textit{request 3} with the prefill chunks~\cite{splitfuse, sarathi} of other requests.
Instead, by scheduling \textit{request 3} earlier, its decoding can be mixed with the prefill chunks of other requests.
Another problem is that, the current systems use the token and request number in the token-batch as the threshold to limit the batching, which limits to batch more tokens together in the decoding dominated token-batches and prevents better utilizing the GPU even when the memory is enough.
Fig.\ref{fig:valley-example} shows the number of tokens at each iteration for a typical industry task (Sec.\ref{subsubsec:case1}) performing with vLLM's best configuration.
It shows that there are "valleys"
at many iterations where the number of tokens are not large enough to saturate the GPU.
There is a demand to reschedule the tasks to make the requests sharing common prefix closer, and enable better mix of decoding tokens with prefill chunks.
There is also a room to enlarge the size of token-batches dominated by the decoding tokens.

\section{Design Methodology}
\label{subsec:design}

\subsection{Overview}
\label{subsec:overview}


\SYS{} has the following insights to address the limitations described in Sec.\ref{subsec:limitations}.

\textbf{1. Explicit prefix identification and sharing.}
Instead of managing the prefix with LRU cache, \SYS{} explicitly identifies the common prefix globally within the large batch ahead-of-time,
which will not miss the opportunity of prefix sharing caused by the implicit cache.
Besides, \SYS{} refactors the prefix tree with a Dynamic Programming algorithm to enlarge the first-level prefix to avoid the system complexity and kernel overhead of the multi-level prefix.

\textbf{2. Grouped scheduling, request reordering and memory-centric token batching.}
\SYS{} schedules the requests at the granularity of a group of requests sharing the common prefix, which makes the prefix sharing convenient and shrinks the lifetime of the prefix's KV memory.
\SYS{} reorders the requests according to the length of the prompt and the estimated decoding length.
As indicated in Fig.\ref{fig:processing-order}, \SYS{} will schedule the requests with larger ratio of decoding length to prompt length with higher priority.
In this way, the longer prompts will be scheduled later and can be better mixed with the earlier decoding tokens.
It also forms the token-batches with the consideration of the KV memory usage, allowing to batch more tokens together to increase the size of token-batches and reduce the "valleys". 


\begin{figure}
    \centering
    \includegraphics[width=0.9\columnwidth]{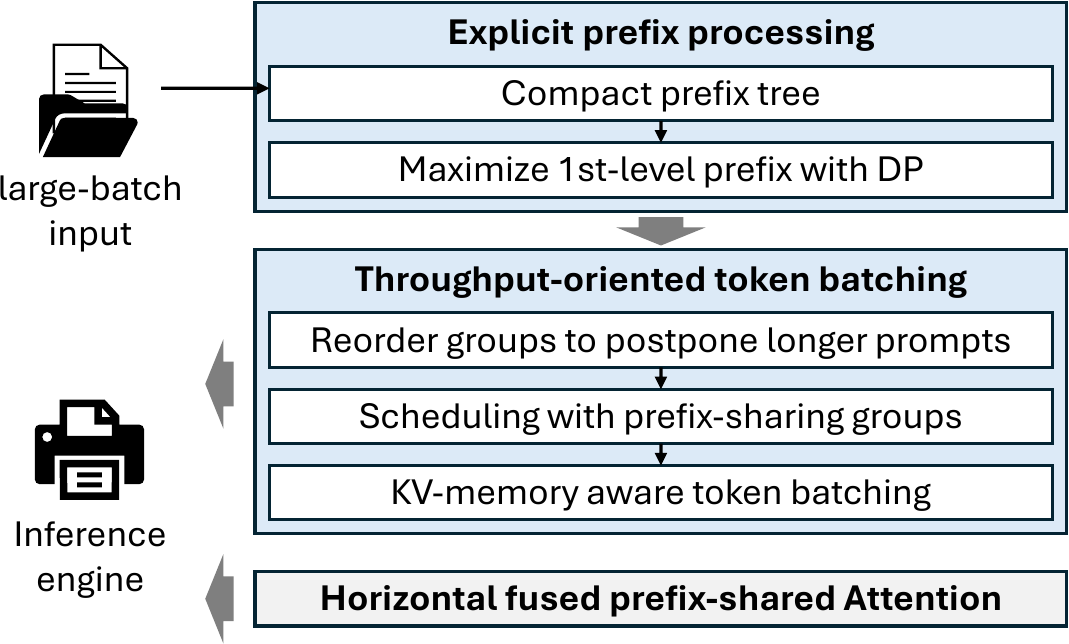}
    \caption{\SYS{} overview.}
    \label{fig:overview}
\end{figure}

Fig.\ref{fig:overview} shows the overview of \SYS{} optimizations.
First, it analyzes the large batch of input prompts and identifies the common prefix explicitly.
This is done before the scheduling of the requests.
It simplifies the multi-level prefix into a single level prefix with the insight that the prefixes of the industry tasks are usually dominated by a long context (Sec.\ref{subsec:scenarios}).
The explicit prefix processing is illustrated in Sec.\ref{subsec:prefix-identification}.
The requests will be organized into groups where each group corresponds to the queries sharing the same prefix.
The group will be the basic unit of task scheduling.
Before scheduling the groups, \SYS{} will also reorder the groups according to the ratio of prefill length and postpone the groups with longer prefill.
It then forms the token-batches with the consideration of the KV memory usage.
This aims to better mix the decoding steps with the prefill chunks to increase the overall token-batch size.
The throughput-oriented scheduling and token-batching optimization is described in Sec.\ref{subsec:token-batching}.

\subsection{Explicit Global KV Reuse Identification}
\label{subsec:prefix-identification}

As discussed in Sec.\ref{subsec:scenarios}, different prompts may have multiple levels of prefixes.
The multi-level prefixes can lead to system challenges and overhead of token batching and fused Attention kernel.
For the token batching, it requires to manage the dependencies of the different levels of prefixes.
For the Attention kernel, it will introduce more GPU kernels of the multi-level reuse.
Instead, \SYS{} compresses the multi-level prefixes into a single level prefix to avoid the above challenges,
based on the insight that the length of the prefixes are usually dominated by a single level (Sec.\ref{subsec:scenarios}).

\begin{figure}
    \centering
    \includegraphics[width=0.95\columnwidth]{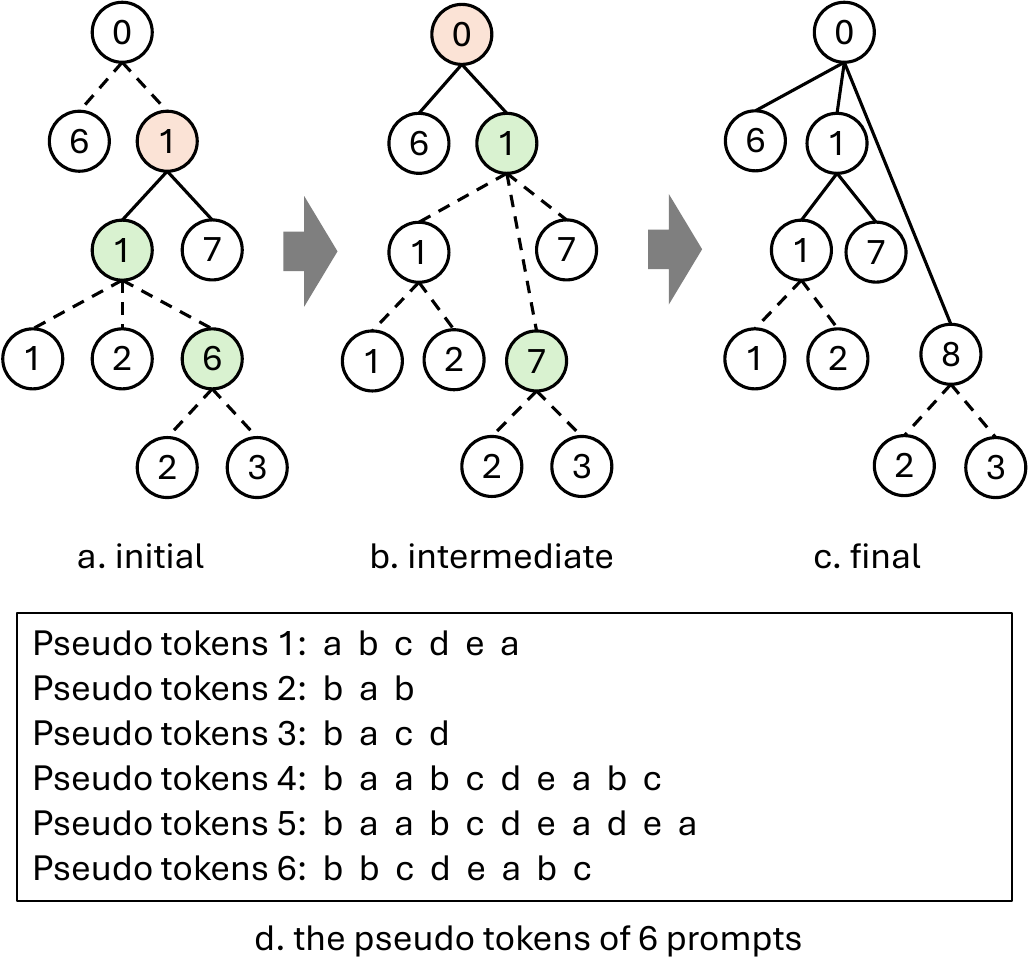}
    \caption{
    The preprocessing to maximize the first level prefix reusing.
    Each \cb{circle} represents a node containing a number of tokens.
    The number in the \cb{circle} is the token number of the node.
    }
    \label{fig:preprocessing}
\end{figure}

\SYS{} explicitly identifies the common prefix among the prompts within the large batch.
It makes use of the compact prefix tree\footnote{The compact prefix tree, also called Radix tree, is the prefix tree that merges the only child with its parent.}
to identify and represent the common prefix between prompts.
However, directly using the first level prefix of the basic prefix tree as the common prefix can lead to suboptimal prefix reusing.
Take the \cb{pseudo} prompts in Fig.\ref{fig:preprocessing}d as an example.
The 6 prompts are organized with the compact prefix tree in Fig.\ref{fig:preprocessing}a, where each node represents several consecutive tokens of the prompt and the tokens in a node is shared by its children.
The right child of the root node (i.e., token 'b') is shared by 5 prompts in total (prompt 2 to 6) according to Fig.\ref{fig:preprocessing}a, thus reusing the first level prefix can save 4 tokens' KV computation.
However, it is easy to infer that sharing the first 8 tokens between prompt 4 and 5 can save 8 tokens' KV computation.
Thus it requires to refactor the naive prefix tree to maximize the achieved first-level token reusing.

\begin{algorithm}
\caption{Dynamic Programming on the prefix tree to maximize the first-level token reusing of node $D$.
}
\label{alg:dp-algorithm}
\begin{algorithmic}[1]
\STATE{$\triangleright$ \cgray{The 1st-level prefixes of a node are its \cb{children}}}
\FUNCTION{MaximizeReuse(D)}
  \IF{D has no children}
    \STATE{\textbf{return}}
  \ENDIF
  \FOR{$child \in D.children$}
    \STATE{MaximizeReuse($child$)}
    \COMMENT{\cgray{Solve sub-problems}}
    \FOR{$gchild \in child.children$}
      \STATE{$gain$ $\gets$ (leaves($gchild$) - 1) $\times$ tokens($gchild$)}
      \STATE{$penalty$ $\gets$ tokens($child$)}
      \IF{$gain$ > $penalty$}
        \STATE{Fork $child$ and merge with $gchild$}
      \ENDIF
    \ENDFOR
  \ENDFOR
\ENDFUNCTION
\end{algorithmic}
\end{algorithm}

We design a Dynamic Programming algorithm on the compact prefix tree to maximize first-level token reusing, shown in Algo.\ref{alg:dp-algorithm}.
The basic insight is that, given a node $D$ to maximize its first-level token reusing, it first solves the sub-problems of all $D$'s children to maximize each child's first-level token reusing (line 6 in Algo.\ref{alg:dp-algorithm}), it then try to merge $D$'s first-level prefix with its children's to see if it can enlarge $D$'s token reusing (line 7-11 in Algo.\ref{alg:dp-algorithm}).
It recursive performs the above procedure until reaching the root note, by calling \textit{MaximizeReuse}(\textit{root}).
Fig.\ref{fig:preprocessing}b and Fig.\ref{fig:preprocessing}c shows an example of this procedure.
In Fig.\ref{fig:preprocessing}b, the left node in level 3 of the tree is forked and merged with it's right child, which enlarges the 1st-level prefix reusing of the right node in level 2.
Similarly, the right node in level 2 is forked and merged with its second child to enlarge the 1st-level prefix reusing of the root node.
After maximizing the first-level prefix, the other levels of prefixes will be expanded during the token batching and will not be the shared context.

\cb{
We have compared the token saving ratio between the original multi-level prefix and the enlarged single prefix.
The saving ratio is calculated according to Eq.\ref{eq:saving-ratio}.
For a dataset of the snippet generation task, the saving ratio of the original multi-level prefixes is about 56\%, and the ratio is about 55\% for the first-level prefix after enlarging (Sec.\ref{subsubsec:case1}).
It means the enlarged first-level prefix only has 1\% loss of the token reuse compared to the multi-level prefix for this typical workload, with the advantage of reducing the complexity and the overhead of the token batching and Attention kernel optimization.
More details will be presented in Sec.\ref{subsubsec:breakdown-prefix-grouping}.
}

\cb{
\textbf{Time and Space Complexity Analysis.}
The complexity of Algo.\ref{alg:dp-algorithm} is $O(n c^2)$, where $n$ is the number of nodes and $c$ is the number of average children.
The algorithm visits every node once and has $O(n)$ for traversal, and iterates over its children and their children for each node and has $O(c^2)$ per node as the worst case.
Thus, the time complexity is $O(n c^2)$.
In real scenarios, $c$ is typically small on average, so this is close to $O(n)$ in practice.
The space complexity is $O(n)$.
The memory overhead introduced by "fork child and merge with grandchild" is very small, as the metadata of the node is the \texttt{token\_id}, which is in \texttt{int64} datatype.
The worst case is that all nodes are fully forked, which means the same prefix of different prompts are stored independently.
In this case, the storage will be equal to the sum of token numbers of all prompts (in \texttt{int64} datatype), which is still small compared to the memory consumption of model weights.
Algo.\ref{alg:dp-algorithm} does not introduce KV residency time as it is computed based on the input \texttt{token\_ids} before processing the tokens.
As a result, the DP algorithm only introduces less than 0.01\% overhead in our evaluation (Sec.\ref{subsubsec:breakdown-prefix-grouping})
}

\cb{
\SYS{} does not completely disable the LRU mechanism.
It explicitly applies global prefix cache to the longest first-level prefix and relies on vLLM’s LRU to manage the possible remaining prefixes (e.g., the second-level sub-prefix) for distinct prompt parts.
}

\subsection{Throughput-oriented Token Batching}
\label{subsec:token-batching}

\subsubsection{Prefix-sharing Group Based Scheduling}

To simplify the KV reuse and reduce the lifetime of the common prefix in the memory, \SYS{} schedules the requests sharing the same prefix all together.
Specifically, the procedure described in Sec.\ref{subsec:prefix-identification} will generate the \textit{prefix-sharing groups} (the collection of requests sharing the same prefix) and \SYS{} will schedule the requests at the granularity of \textit{prefix-sharing group} rather than each single request.
In this way, the requests sharing the same prefix can start at the same time and the memory of the common prefix can be released as soon as the group is completed.
This differs from the LRU cache based prefix management and can make sure all the common prefix can be reused the best and the memory will not be retained unnecessarily.

To realize the \textit{prefix-sharing group} based scheduling, \SYS{} maintains three queues of the to-be-batched tokens: common prefix queue, distinct prompt queue, and decoding queue.
At each iteration, \SYS{} will form the token-batch by fetching the tokens from the three queues in the order of decoding queue, distinct prompt queue, and common prefix queue
(one constraint is that the a common prefix should be fetched and processed before its corresponding distinct prompts).
On the one hand, fetching the decoding tokens before prefills can help to better mix the two types of tokens together in the token-batch~\cite{splitfuse}.
On the other hand, fetching the decoding and distinct prompt tokens before the prefix tokens can make sure the active requests (i.e., the request that has been partially processed) are scheduled before the inactive requests, which finishes the active requests earlier and releases their memory earlier.

\subsubsection{Request Groups Reordering}
\label{subsubsec:group-reordering}

\SYS{} reorders the input requests to schedule the requests according to the ratio of prompt length to decoding length ($R$).
We define the ratio $R$ as:
\begin{equation}
R = \frac{L_{decode}}{L_{prefill}}
\label{eq:order-metric}
\end{equation}
In Eq.\ref{eq:order-metric}, $L_{prefill}$ is the number of prefill tokens and $L_{decode}$ is the length of the output tokens.
The requests with larger $R$ will be scheduled earlier. 
This helps to issue the requests with long decoding steps earlier and 
make the latter prompts' length larger and thus can be better mixed with the previous decoding tokens to enlarge the token-batch size (example in Fig.\ref{fig:processing-order}).
The challenge is that the exact number of output tokens is not known before finishing the execution.
We profiled the dataset of several typical industry tasks (ads, caption generation, offline ranking, etc.) and observed that the distribution of output length is relatively concentrated for the same kind of task,
which means the effect of the output length in Eq.\ref{eq:order-metric} is less significant for the same kind of task as it does not vary a lot.
We thus make an approximation by normalizing $L_{decode}$ into 1 in Eq.\ref{eq:order-metric} for all requests uniformly.
As we schedule the requests in the unit of \textit{prefix-sharing group}, the ratio of the group (after normalization) becomes: 
\begin{equation}
R_{group} = \frac{1}{L_{prefix} + \sum L_{distinct}}
\label{eq:group-ratio}
\end{equation}
The request groups with larger $R_{group}$ will be scheduled with higher priority.

\begin{figure}[]
    \centering

	\subfloat[ Output length distribution of Ads title rewrite.]{
		\begin{minipage}[b]{1\columnwidth}
  	 	\includegraphics[width=1\textwidth]{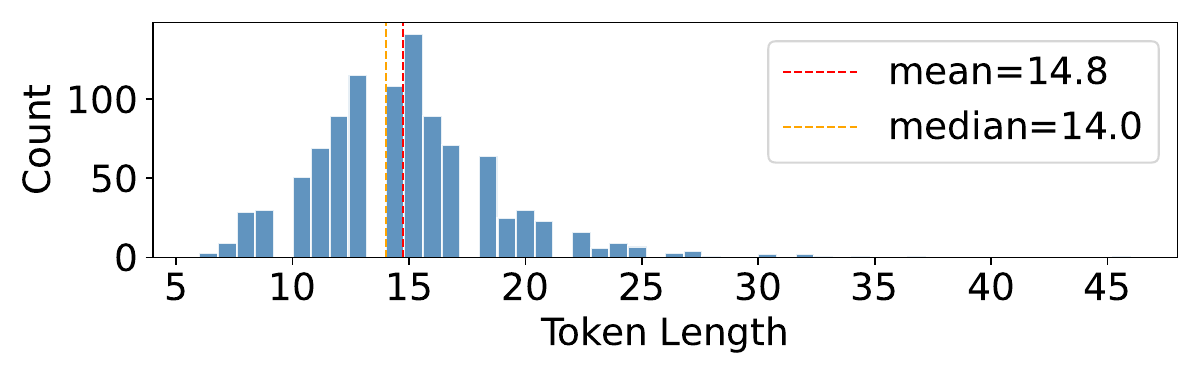}
		\end{minipage}
	    \label{fig:out-dist-ads}
	} \\

     \subfloat[ Output length distribution of caption generation.]{
		\begin{minipage}[b]{1\columnwidth}
			\includegraphics[width=1\textwidth]{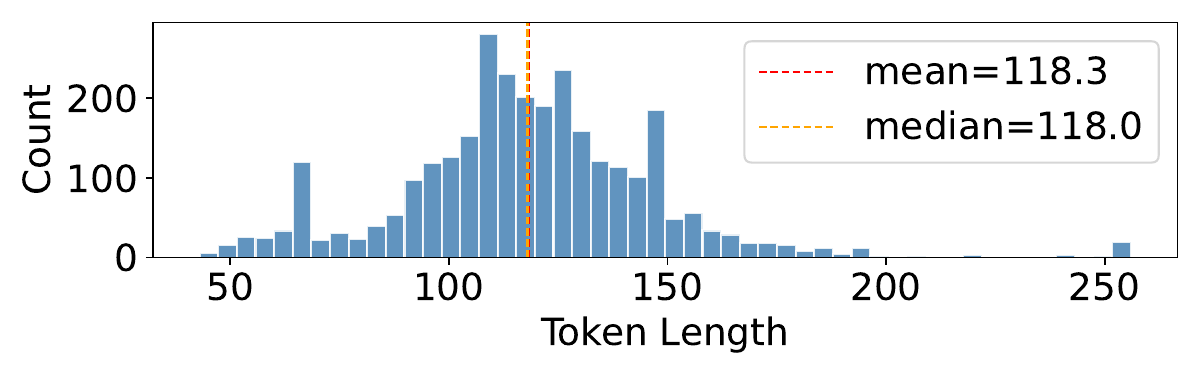}
		\end{minipage}
		\label{fig:out-dist-caption}
	} \\
    \caption{\cb{Generation output length distribution of two typical real-industry workloads.}}
    \label{fig:industry-output-len}
\end{figure}

\cb{
Fig.\ref{fig:industry-output-len} shows the output length distributions of two representative industry workloads.
It shows that the the output lengths are in a very narrow range, 
which motivates our choice of normalizing decoding length.
}

\subsubsection{Memory-centric Token Batching to Saturate GPU}
\label{subsubsec:memory-centric-batching}

Another limiter of the size of token-batches is that, the existing works~\cite{vllm} use a fixed request number to limit the token batching, i.e., the request number cannot exceed a fixed threshold within a token-batch.
For a token-batch dominated by the decoding tokens, it is easy to reach the upper bound of the request number while still have not achieved a large number of the total tokens.
For example, vLLM sets the threshold of request number as 256 by default, if a token-batch already has 256 decoding tokens, it will have no chance to add more prefill chunks into this token-batch even if there is still enough memory to hold the KV memory of the prefill chunks.
Note that purely increasing this threshold does not solve the problem in practice (noted by the vLLM community\footnote{
https://github.com/vllm-project/vllm/issues/6801}),
as directly excessively enlarging batch size may cause the engine to have more frequent memory swaps and recomputation due to the GPU memory can be not enough. 


The target of the token-batching is to enlarge the number of tokens in the batch under the constraint of GPU memory size.
Thus there are two main factors that matters for the token-batching procedure: whether the number of tokens is large enough in the batch to saturate the GPU, indicating the current status of the token-batch; and whether the remaining memory is enough to accommodate more prefill chunks, indicating if the status can be improved.
With this insight, \SYS{} uses the KV memory itself and the predetermined chunk size $S_{chunk}$ as the limiter rather than using the indirect limiter of the request number.

Specifically, \SYS{} predefines a size as the upperbound of the KV memory size ($M_{threshold}$).
When deciding the token-batching at each iteration, it will only check whether adding a prefill chunk will exceed $M_{threshold}$ or not, without considering the number of requests.
The prefill chunk will be added into the token-batch if the remaining GPU memory capacity allows it.
Note that we also use a fixed number to limit the per-batch token number for token-batching decision (noted as the chunk size $S_{chunk}$), which aligns with that in vLLM and helps to distribute the token numbers in different iterations more even.

\subsection{\cb{End-to-end System Integration}}

\cb{
The integration of \SYS{} into other frameworks is quite easy. \SYS{} comprises two pluggable modules: (i) a token-batching scheduler and (ii) an attention backend, that can be added alongside existing components in common inference frameworks (e.g., vLLM, SGLang) without altering user-facing APIs or the serving loop.
Token-batching scheduler is orthogonal to the framework’s existing continuous batching/prefill–decode pipeline. We add lightweight per-request metadata to indicate shared prefix vs. distinct part. The scheduler ensures the common prefix across requests is computed once and reused, reducing redundant calculation in both attention (KV-cache construction) and non-attention compute (e.g., projection and FFN GEMMs) during prefill and decoding.
Attention backend is a drop-in backend with full fallback. The backend presents the same public interface as the framework’s attention module and simply registers as an additional backend. If shapes/hardware are unsupported, it falls back to the standard kernel, ensuring compatibility and graceful degradation.
}

\begin{figure}[]
    \centering
    \subfloat[Llama 3.1 8B throughput on A100, \#requests = 6400]{
		\begin{minipage}[b]{.95\columnwidth}
            \includegraphics[width=1\textwidth]{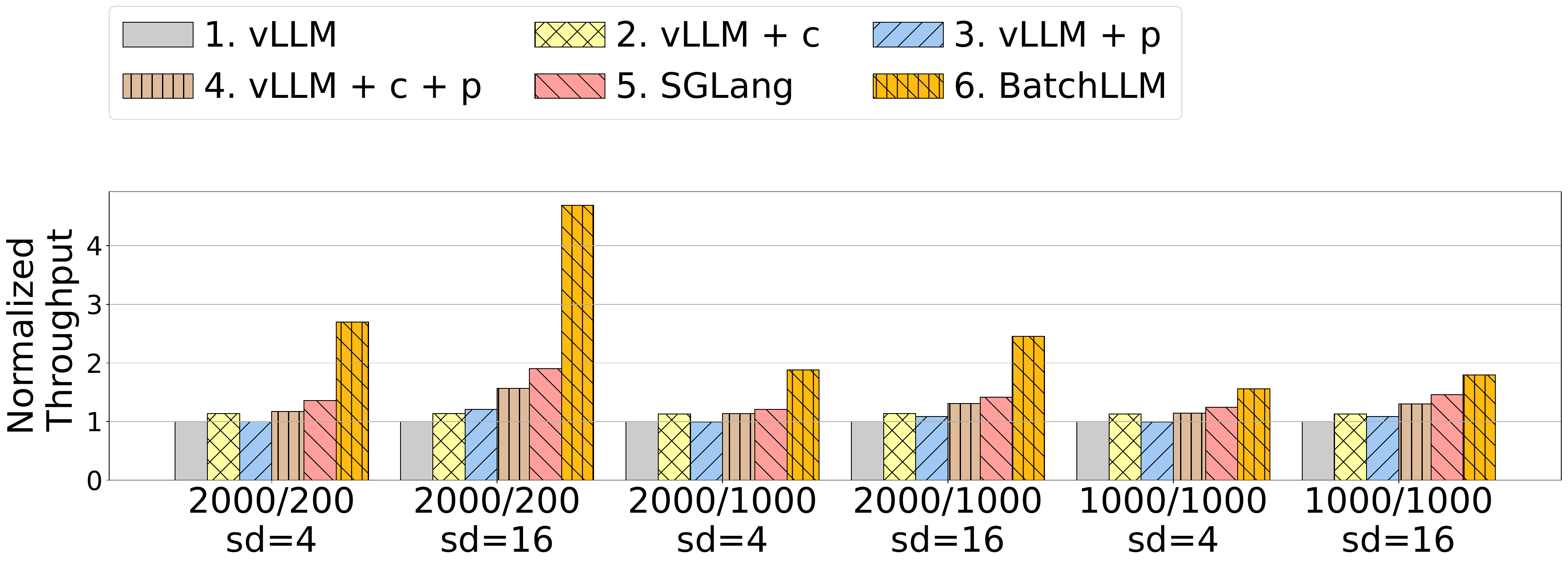}
		\end{minipage}
		\label{fig:llama-3-1-8b-a100-micro}
	} \\
	\subfloat[Qwen 2.5 14B throughput on A100, \#requests = 6400]{
		\begin{minipage}[b]{.95\columnwidth}
            \includegraphics[width=1\textwidth]{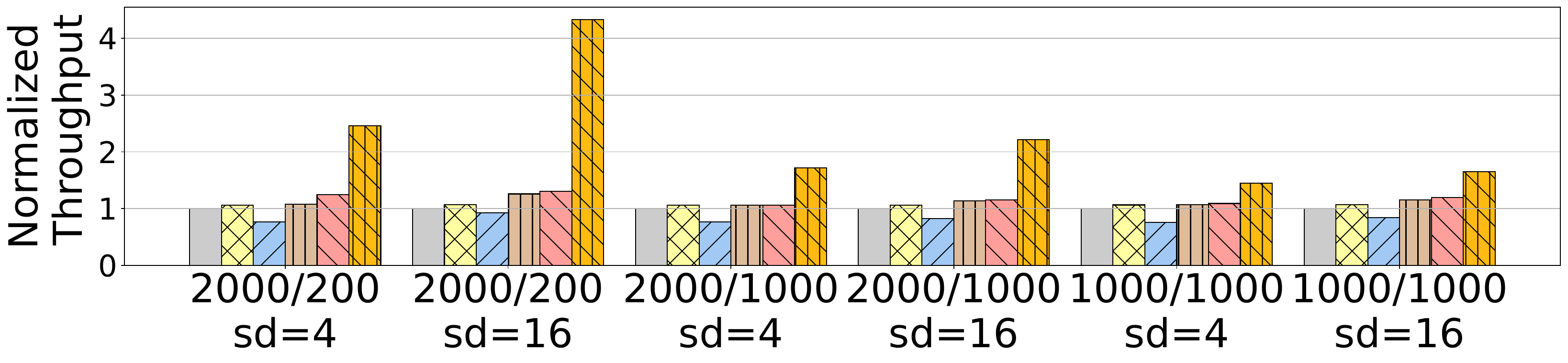}
		\end{minipage}
	    \label{fig:qwen-2-5-14b-a100-micro}
	} \\  
	\subfloat[Llama 3.1 70B throughput, on A100, \#requests = 800]{
		\begin{minipage}[b]{.95\columnwidth}
  	 	\includegraphics[width=1\textwidth]{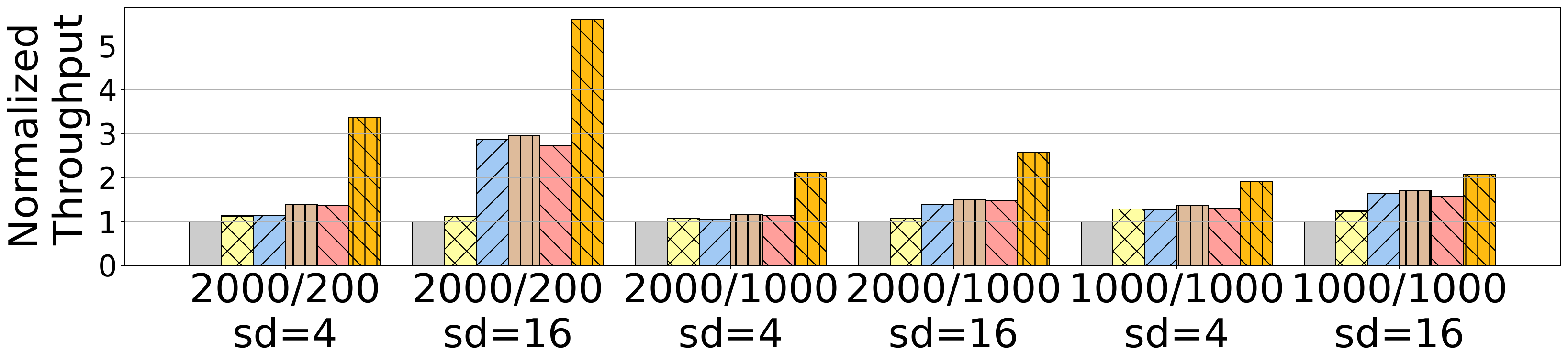}
		\end{minipage}
	    \label{fig:llama-3-1-70b-a100-micro}
	} \\
    \subfloat[Llama 3.1 70B throughput, on A100-TP2, \#requests = 800]{
		\begin{minipage}[b]{.95\columnwidth}
  	 	\includegraphics[width=1\textwidth]{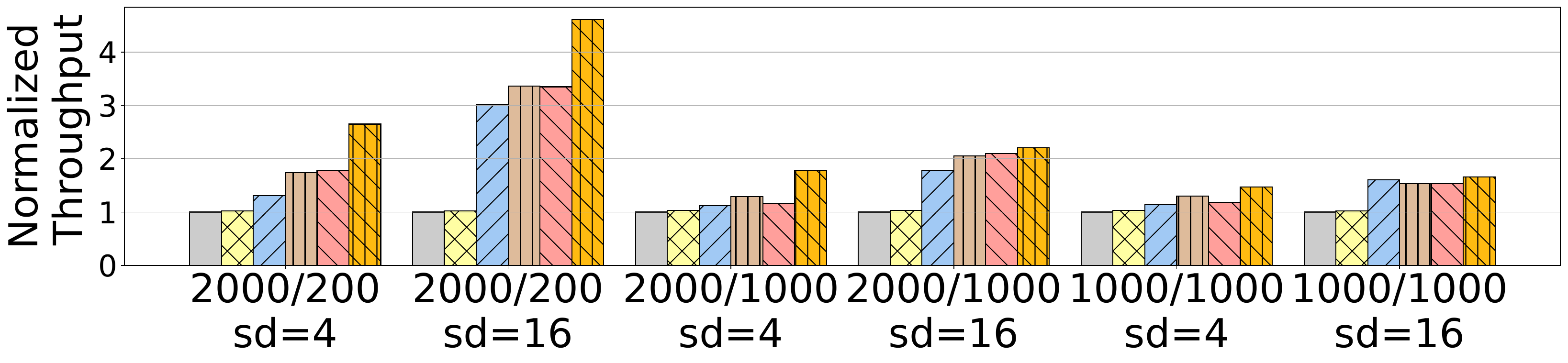}
		\end{minipage}
	    \label{fig:llama-3-1-70b-a100tp2-micro}
	} \\
    \subfloat[Llama 3.1 8B throughput with Poisson perturbed lengths on A100, \#requests = 6400]{
		\begin{minipage}[b]{.95\columnwidth}
  	 	\includegraphics[width=1\textwidth]{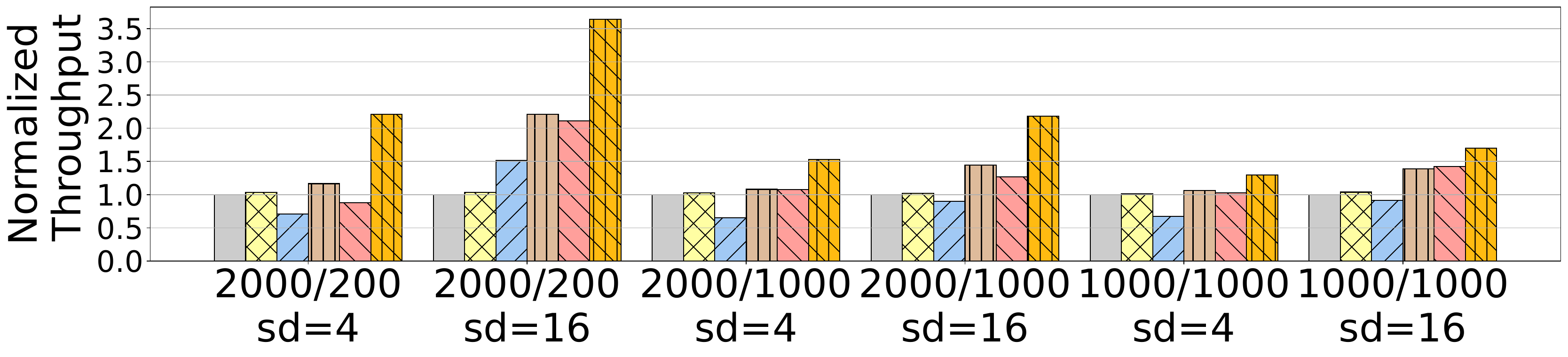}
		\end{minipage}
	    \label{fig:llama-3-1-70b-a100Poisson-micro}
	} \\
	\subfloat[Llama 3.1 8B throughput, on MI200, \#requests = 6400]{
		\begin{minipage}[b]{.95\columnwidth}
  	 	\includegraphics[width=1\textwidth]{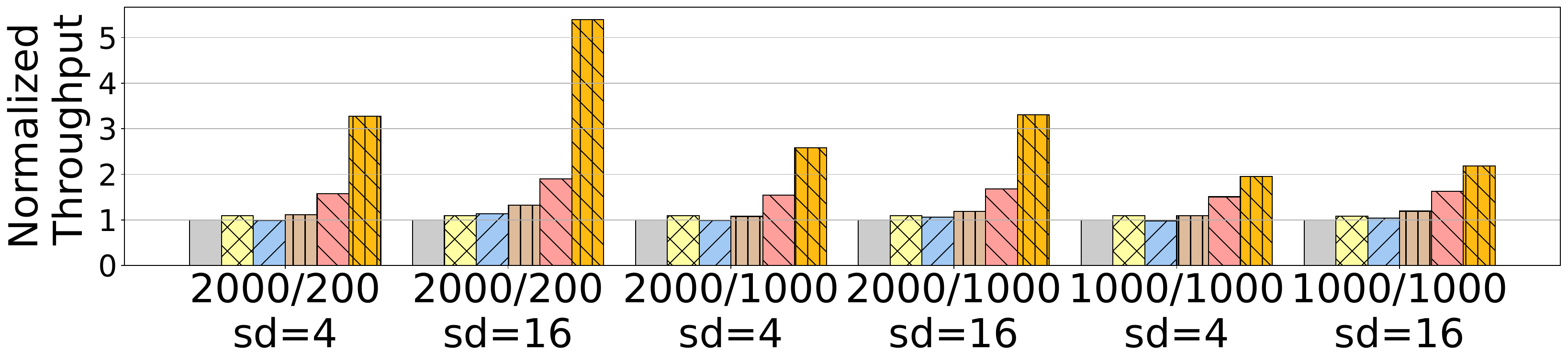}
		\end{minipage}
	    \label{fig:llama-3-1-8b-mi200-micro}
	} \\
	\subfloat[Qwen 2.5 14B throughput, on MI200, \#requests = 6400]{
		\begin{minipage}[b]{.95\columnwidth}
  	 	\includegraphics[width=1\textwidth]{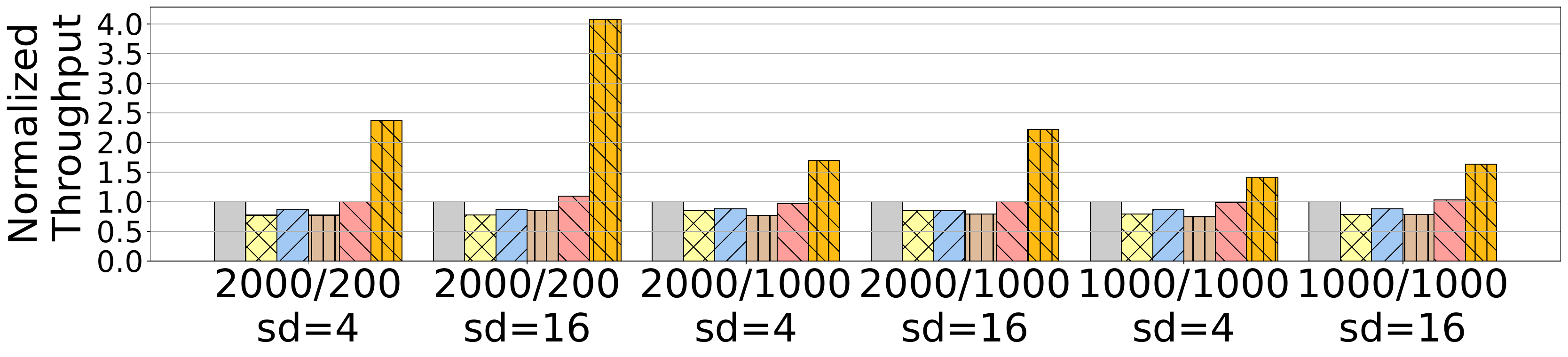}
		\end{minipage}
	    \label{fig:qwen-2-5-14b-mi200-micro}
	} \\
    \caption{Microbenchmark evaluation. The setting \textit{m/n} (like \textit{2000/200}) indicates the length of shared prefix/non-shared context,
    \textit{sd} means \textit{sharing degree}. For subfigure (e), lengths are perturbed around base values using a Poisson distribution.
    The vLLM setting with '+ p' ('+ c') means \textit{prefix-caching} (\textit{chunked-prefill}) enabled.
    }
    \label{fig:eval-microbenchmark}
\end{figure}

\section{Evaluation}
\label{sec:evaluation}


\subsection{Setup}
\label{subsec:setup}

\textbf{Baseline Specification.}
We compare \SYS{} with vLLM and SGLang, two state-of-the-art LLM inference engines with prefix sharing, for the end-to-end comparison to demonstrate the efficacy (Sec.\ref{subsec:e2e}).
We integrate all our design in vLLM v0.6.4 and the baseline for comparison is also vLLM v0.6.4 (except as noted).

We enable the \textit{prefix-caching} and \textit{chunked-prefill} in vLLM baseline for the end-to-end comparison in Sec.\ref{subsec:e2e}.
The \textit{prefix-caching} uses the implicit LRU-based caching policy.
We use 2048 as the token-batch size for chunked-prefill of vLLM baseline to maximize its throughput.

\textbf{Hardware and System Specification.}
We conduct our experiments on both NVIDIA A100 80GB GPUs (both single GPU and 2-GPUs) and AMD MI200 GPU.
For multi-GPU experiments on NVIDIA A100, we utilize tensor parallelism (TP) with size of 2 (denoted as A100-TP2).
The CUDA version on the NVIDIA platform is 12.1.
The ROCm version on the AMD platform is 6.1.
The CPU is AMD EPYC 7V12 CPU @2.45GHz.
The operation system is Ubuntu 20.04.

\textbf{Models and Dataset.}
We evaluate \SYS{} and the baselines with Llama-3.1-8B~\cite{llama3modelcard}, Qwen-2.5-14B~\cite{qwen25} and Llama-3.1-70B (4-bit GPTQ),
with a set of input/output with different data distributions,
and Qwen-2.5-7B for a typical industry task.
As for the basic evaluation, we use three different settings of the length of common prefix and distinct prompt.
The length of prefix/distinct are 2000/200, 2000/1000 and 1000/1000 respectively, while the length of generated tokens is 100.
We evaluate the three settings under different share-degrees.
The share-degree means the number of distinct prompts sharing the same prefix. 
Note that all requests are randomly shuffled.
The above input/output settings are very representative as they come from the real industry workloads running on thousands of GPUs.
We use a uniform length to avoid analysis inconvenience caused by different lengths.
Additionally, to assess performance under more dynamic lengths, we conduct experiments on Llama-3.1-8B (A100, TP=1) where the lengths of the shared prefix, distinct context, and generated tokens are perturbed around base values using a Poisson distribution. These experiments help evaluate the robustness of \SYS{} and baselines to variations in sequence lengths.
The microbenchmark evaluation is in Sec.\ref{subsec:microbenchmark}.
As for the industry tasks, we conduct the evaluation on a typical scenario in web search engines, which is the snippet generation task that generates the snippet of a document for the user query.
We conduct the breakdown experiments in Sec.\ref{subsec:breakdown} to demonstrate the efficacy of each technique.

\subsection{End to End Comparison}
\label{subsec:e2e}

\begin{figure}[]
    \centering
    
    \captionsetup[minipage]{labelformat=empty}
    \subfloat{
        \begin{minipage}[b]{1\columnwidth}
            \includegraphics[width=1\textwidth]{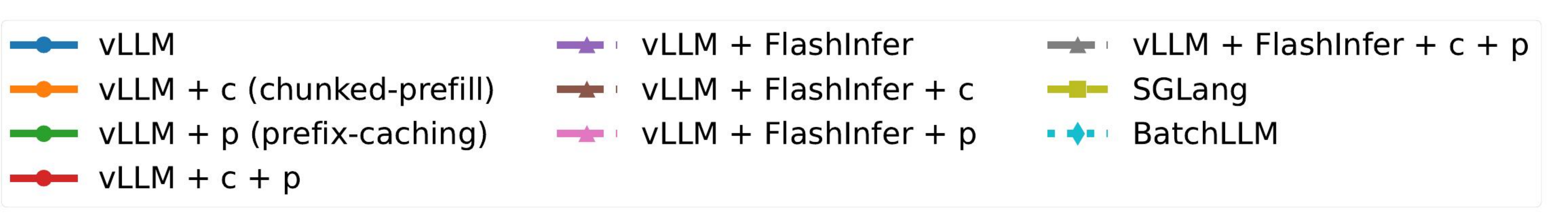}	
        \end{minipage}
        \label{fig:placeholder}
    } \\
    \subfloat{
        \begin{minipage}[b]{1\columnwidth}
            \includegraphics[width=1\textwidth]
            {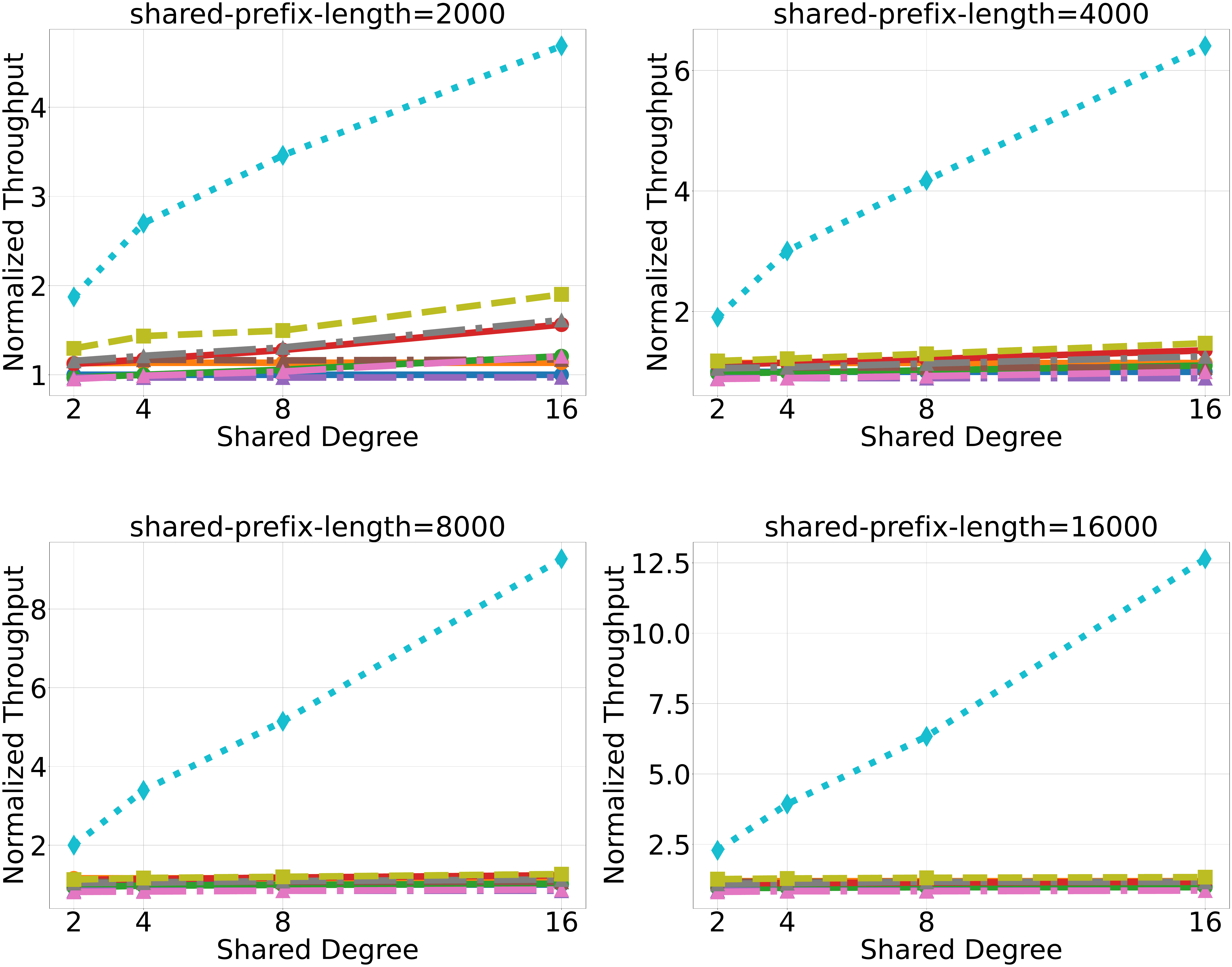}
        \end{minipage}
        \label{fig:placeholder2}
    } \\
    
    \caption{Microbenchmark evaluation of different sharing-degrees and different shared-prefix lengths on A100. Given 6400 requests.
    This evaluation is based on Llama-3.1-8B-Instruct, while the length of non-shared context is 200, and the generated length is 100.}
    \label{fig:diff-sd-diff-prefix-len}
\end{figure}

\subsubsection{Basic Evaluation}
\label{subsec:microbenchmark}


Fig.\ref{fig:eval-microbenchmark} shows the end-to-end throughput comparison between \SYS{} and the baselines with different shared-prefix/non-shared context settings.
The results show that \SYS{} outperforms the vLLM and SGLang baseline under different configurations, for different models on different GPU platforms with different input data distributions.
Specifically, when comparing with \textit{vLLM + prefix-caching + chunked-prefill}, which performs the best across all vLLM's configurations, \SYS{} can get 1.4-3$\times$ speedup on A100 and 1.8-4.1$\times$ speedup on MI200. \SYS{} also achieves better performance than SGLang, with 1.2-2.5$\times$ speedup on A100 and 1.3-2.8$\times$ speedup on MI200. The acceleration is stable from 7B to 70B models.

\textbf{The impact of sharing-degree and shared prefix}.
We also compare the throughput of different settings after changing the sharing-degree and the length of shared prefix.
Fig.\ref{fig:diff-sd-diff-prefix-len} shows that the speedup becomes more significant when more requests share the same prefix and the portion of the common prefix becomes larger, which contributes to a bigger reuse ratio, and \SYS{} can save more computation and memory for this case. 
With lower sharing-degree and shorter common prefix (like \textit{sd} is 2 and the length of prefix is 2000), the acceleration of \SYS{} is $1.7\times$/$1.5\times$, comparing with the best setting of vLLM and SGLang. When the length of shared prefix is 16000, and the share degree is 16, the best performance of vLLM and SGLang are 0.49 and 0.56, while \SYS{} achieves $10.8\times$/$9.5\times$ speedup.

\textbf{Comparison between vLLM + request sorting and \SYS{}.} 
We perform a preprocess for the setting of \textit{vLLM + chunked-prefill + prefix-cahcing}: use \textit{python.sorted()} to gather all requests with the same prefix together\footnote{
Here the length of shared prefix/non-shared context/generated tokens are 2000/200/100, while the sharing degree is 16.
The model is Llama-3.1-8B. And the GPU is NVIDIA A100.}. 
This can help vLLM to cache the tokens better.
After this preprocess, vLLM's throughput is boosted from 6 to 13.02, while the throughput of \SYS{} is 18.
One reason is that vLLM cannot simplify the attention calculation when there are more than 1 common prefix in the batch every step of its inference, while \SYS{} could handle it.
Besides, \SYS{} has a better token-batching strategy than vLLM.

\cb{
\textbf{Comparison with the latest prefix-caching advances by early 2026.}
\SYS{} is implemented based on an old vLLM version, v0.6.4.
The latest vLLM version has integrated many of the most recent prefix-sharing technologies from the community, and we make a comparison against it.
We evaluate the industry workload in this paper with the latest vLLM (the vLLM's github commit by 1/16/2026).
Its throughput on A100 is 6.57 for this workload with its latest prefix-caching optimization, while \SYS{} achieves 8.67 even with a very old vLLM as the implementation base.
}

Overall, the performance improvement of \SYS{} comes from the better KV reuse with the explicit prefix identification, the token-batching to better mix the decoding with prefill chunks together, and the better Attention kernels,
which we will analyze through the breakdown in Sec.\ref{subsec:breakdown}.

\begin{table}[]
\center
\scriptsize
\caption{Industry workload evaluation.}
\label{tab:realworld_workload}
\begin{tabular}{@{}l|c|cccc@{}}
\toprule
Settings                              & Throughtput, A100 & Throughput, MI200   \\ \midrule
vLLM                               &  5.45 & 2.75 \\ 
\hspace{1em} + c (\textbf{c}hunked-prefill)                      &  5.48 & 2.84 \\
\hspace{1em} + p (\textbf{p}refix-caching)                 &  6.2  & 3.27   \\
\hspace{1em} + c + p      &  6.71 & 3.36 \\ \midrule
\SYS{} &  \textbf{8.67($1.30\times$)}  & \textbf{4.24($1.26\times$)}  \\ \bottomrule
\end{tabular}
\end{table}

\begin{figure}[]
    \centering
    \includegraphics[width=1.0\columnwidth]{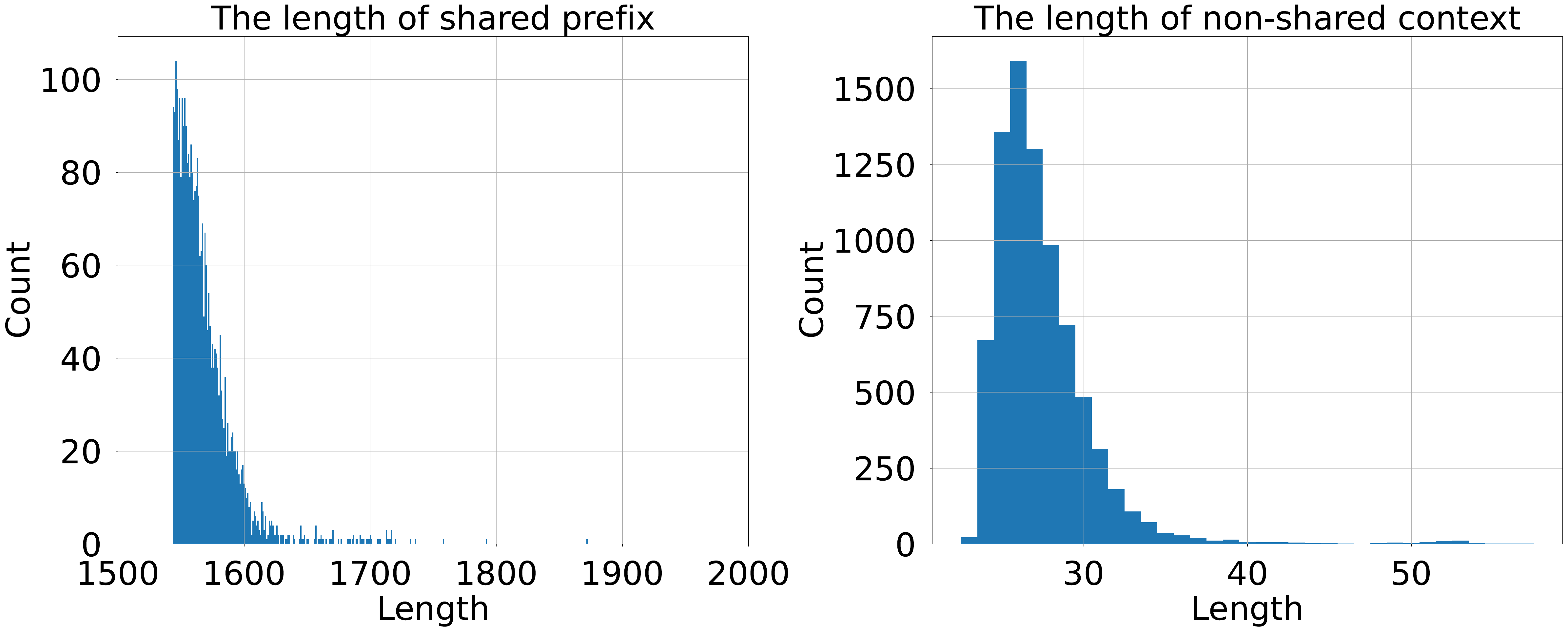}
    \caption{The length distribution of shared prefix/non-shared context in the dataset of the industry workload.}
    \label{fig:distribution-cap}
\end{figure}

\subsubsection{Industry Workload Evaluation}
\label{subsubsec:case1}

This industry workload is similar to the snippet generation task described in the recent work~\cite{large-search-model}, which is also described in Sec.\ref{subsec:scenarios}.
The original input has two levels of prefixes: the global shared instruction to extract the information from the document-query pair, and the shared document between different groups of queries.
The first-level prefix is very short.
With the optimization described in Sec.\ref{subsec:prefix-identification}, the first-level prefixes are enlarged (i.e., some of the second-level prefix of the document is merged into the first-level prefix).
We have analyzed the token saving ratio according to Eq.\ref{eq:saving-ratio}.
It shows that the saving ratio are nearly the same between the multi-level and the enlarged single level prefixes, 56\% for the former and 55\% for the latter.
Besides, the DP algorithm only introduces less than 0.01\% overhead.

The data distribution of this workload is shown in Fig.\ref{fig:distribution-cap}.
The average share-degree (i.e., the number of the distinct prompt that share the same prefix) is about 3, the average common prefix length is about 1570 tokens, and the average distinct prompt length is about 30 tokens.
We sample 8000 queries for this case study.

We conduct this experiment on the NVIDIA A100 GPU and AMD MI200 GPU.
The model is a fine-tuned Qwen-2.5-7B model.
Table.\ref{tab:realworld_workload} shows the effectiveness of \SYS{} of this workload, with about 1.3$\times$/1.26$\times$ speedup over the best configuration of vLLM.
\SYS{} better reuses the KV context than the vLLM baseline with the optimization in Sec.\ref{subsec:prefix-identification}, for which we have the breakdown analysis in Sec.\ref{subsubsec:breakdown-prefix-grouping}.
\SYS{} also better mixes the decoding tokens with prefill tokens to increase the size of token-batches with the optimization in Sec.\ref{subsec:token-batching},
for which we have the breakdown analyze in Sec.\ref{subsubsec:breakdown-reordering}.
It also benefits from the optimized Attention kernel described later.
\cb{During the execution, it does not have KV swaps/recomputations in \SYS{} when processing the industry workload above.
The peak KV usage is about 50\% for this workload.}




\subsubsection{\cb{Heavy-tailed decoding with varied output lengths.}}

\cb{
We evaluate heavy-tailed decoding to demonstrate \SYS{}'s efficacy for varied output lengths, using a Pareto-distributed output length configuration.
}

\cb{
We generate a workload of 320 requests sampled from a Pareto distribution (shape parameter $\alpha = 1.5$) bounded between 10 and 512 tokens.
This distribution yields a mean generation length of 51.2 tokens.
However, the heavy-tail nature of the workload is best illustrated by its percentiles: while the median ($p50$) is just 25 tokens, the 99th percentile ($p99$) reaches 509 tokens, resulting in an extreme $p99/p50$ ratio of 20.3$\times$.
The evaluation is conducted using the Qwen 2.5 7B model at FP16 precision.
To stress-test the prefix-sharing capabilities of our system, the 320 requests are divided into 20 groups.
Each group consists of 16 requests that share a common, lengthy prefix of 2000 tokens, appended with a distinct prompt of 200 tokens (totaling an input length of 2200 tokens per request).
We enable chunked prefill with a maximum of 2048 batched tokens and configure a KV cache block size of 64.
}

\cb{
We compare \SYS{} with vLLM.
Tab.\ref{tab:heavy_tail_throughput} summarizes the end-to-end throughput under the heavy-tail workload.
\SYS{} achieves a substantial 3.2$\times$ speedup over the baseline, increasing throughput from 5.26 req/s to 17.02 req/s. This performance gain demonstrates that \SYS{}'s context sharing mechanism is highly resilient to the decoding inefficiencies traditionally caused by heavy-tail distributions.
}

\begin{table}[]
    \center
    \scriptsize
    \caption{\cb{End-to-end throughput under the heavy-tail workload.}}
    \label{tab:heavy_tail_throughput}
    \begin{tabular}{ccc}
        \toprule
        Configuration & Throughput (req/s) & Speedup \\
        \midrule
        vLLM & 5.26  & 1.0$\times$ \\
        \SYS{} & 17.02 & \textbf{3.2$\times$} \\
        \bottomrule
    \end{tabular}
\end{table}

\subsection{Breakdown Analysis}
\label{subsec:breakdown}

\cb{
To better isolate the performance improvements of \SYS{}, we provide the breakdown analysis to evaluate the individual effects of our core techniques: prefix grouping, request reordering (token-batching), and kernel fusion.
}

\subsubsection{Effect of Prefix Grouping}
\label{subsubsec:breakdown-prefix-grouping}

\cb{
\textbf{Algorithm Effectiveness on Multi-level Prefix Inputs.} 
To demonstrate the effectiveness of our proposed prefix identification algorithm, we conduct an ablation study comparing it against a baseline sorting-based heuristic approach for multi-level prefix inputs.
The heuristic method relies on lexicographical sorting and greedy scanning, grouping requests only if their longest common prefix exceeds a rigid 50\% length threshold relative to the shortest member.
In contrast, our approach automatically discovers shared prefixes of arbitrary lengths without rigid ratio constraints, merging paths based on a dynamic gain formula.
}

\cb{
We construct a synthetic dataset to evaluate edge cases where prompts share significant context but fall short of heuristic thresholds.
Each prompt consists of $1000$ tokens formulated as: $group\_prefix$ + $sub\_prefix$ + $random\_suffix$. We generate $6400$ prompts divided into $50$ global groups.
Each group contains $64$ subcategories, with $2$ prompts per subcategory.
We test two configurations: Setting A uses a $490$-token $group\_prefix$ and an $11$-token $sub\_prefix$, while Setting B uses a $400$-token $group\_prefix$ and a $101$-token $sub\_prefix$. 
}

\cb{
In both settings, the prefix ratio within the same subcategory is $>50\%$ (e.g., $(490+11)/1000 = 50.1\%$), but the prefix ratio across different subcategories within the same global group falls strictly below $50\%$ (e.g., $490/1000 = 49\%$).
}

\begin{table}[h]
\centering
\scriptsize
\caption{\cb{Performance comparison of prefix grouping algorithms across $6400$ requests. The baseline executes without context sharing (CS) or \SYS{} optimizations.}}
\begin{tabular}{@{}lcc@{}}
\toprule
\textbf{Configuration} & \textbf{Saving Ratio} & \textbf{Throughput (req/s)} \\ \midrule
\multicolumn{3}{c}{\textit{Setting A ($group\_prefix=490$)}} \\ \midrule
Baseline (No CS) & 0.0\% & 10.7 \\
Heuristic CSGroup & $\sim$25.0\% & 12.4 \\
\SYS{} & \textbf{48.6\%} & \textbf{15.9} \\ \midrule
\multicolumn{3}{c}{\textit{Setting B ($group\_prefix=400$)}} \\ \midrule
Baseline (No CS) & 0.0\% & 10.7 \\
Heuristic CSGroup & $\sim$25.0\% & 12.3 \\
\SYS{} & \textbf{39.7\%} & \textbf{14.5} \\ \bottomrule
\end{tabular}
\label{table:grouping-ablation}
\end{table}

\cb{
As shown in Table~\ref{table:grouping-ablation}, the heuristic approach is bottlenecked by its 50\% threshold. It successfully groups prompts within the same subcategory (forming $3200$ groups of $2$ members), yielding a limited token saving ratio of $\sim$25\% and throughput of $\sim$12.4 prompts/s. 
}

\cb{
Conversely, our longest prefix identification algorithm completely bypasses this limitation, identifying the global $group\_prefix$ and clustering all $128$ members of a group together (forming $50$ groups).
In Setting A, \SYS{} achieves a 48.6\% token saving ratio and a throughput of 15.9 prompts/s, delivering a 1.28$\times$ speedup over the heuristic method and a 1.49$\times$ speedup over the baseline.
Furthermore, \SYS{} grouping closely matches theoretical ground truth grouping limits, proving its robustness in capturing complex, multi-level prefix hierarchies.
}

\cb{
\textbf{KV Cache Reusing Effect.} 
To show that this grouping optimizes the KV reuse, we evaluate the throughput and the token saving ratio in both microbenchmarks and the industry task. 
Note that all samples in the the industry task are not sorted too.
}

\begin{figure}[]
    \centering
    \includegraphics[width=1.0\columnwidth]{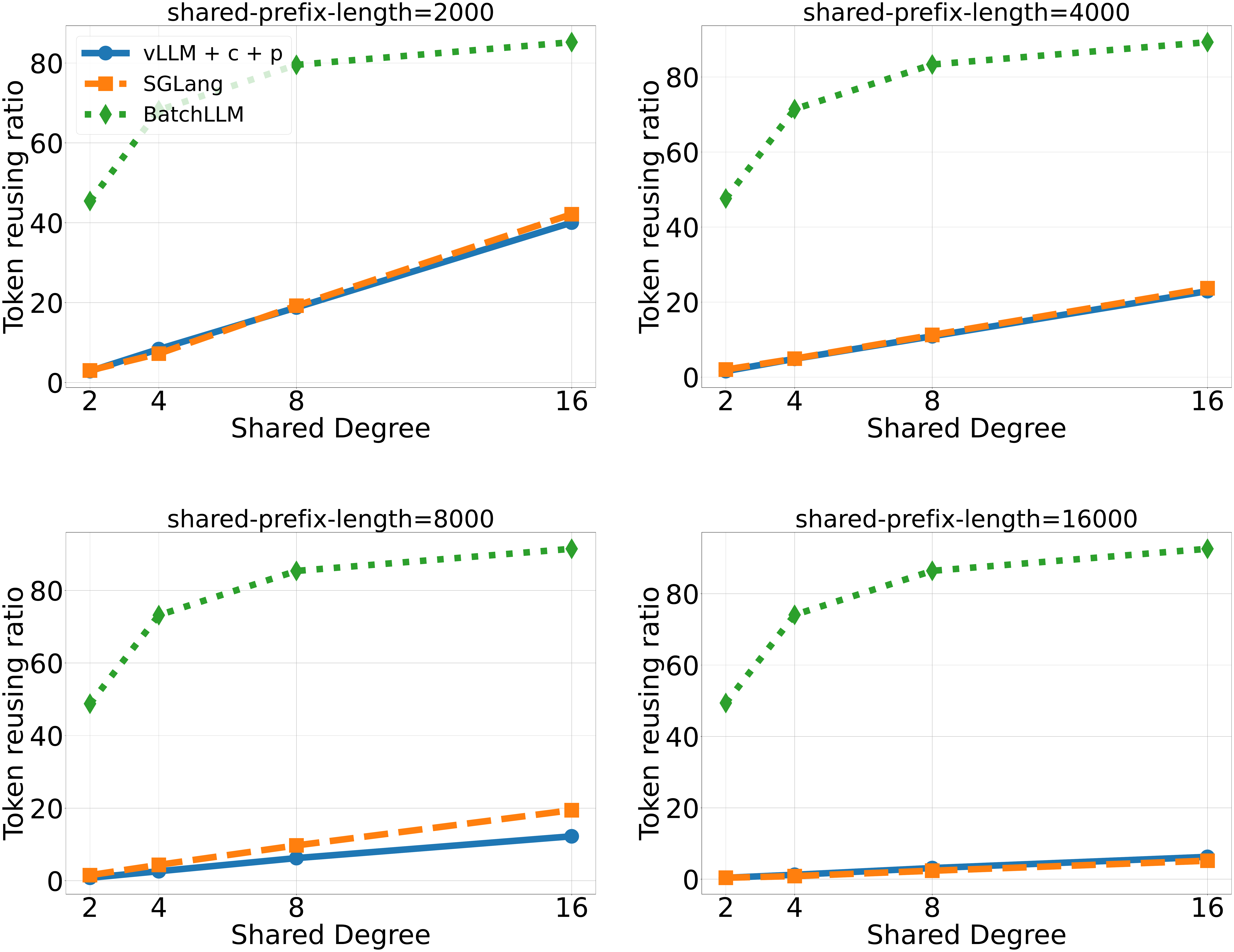}
    \caption{The token saving (or reusing) ratio of different sharing-degrees and different shared-prefix lengths on A100, based on the microbenchmark. This evaluation is based on Llama-3.1-8B-Instruct, while the length of non-shared context is 200, and the generated length is 100.
    }
    \label{fig:token-reusing-ratio}
\end{figure}

As shown in  Fig.\ref{fig:token-reusing-ratio}, we can see the token saving ratio is improved in \SYS{} across different settings. And it becomes higher when the sharing-degree is larger.
Specifically, in the microbenchmark setting where length of shared prefix is 2000 and the \textit{sharing-degree} is 16, \SYS{} achieves 85.2\% token saving ratio, the upper bound in this scenario, comparing to the 40.1\% token saving ratio in \textit{vLLM + chunked-prefill + prefix-caching}.
With the increase of the requests number and the length of prefix, it can be expected that the implicit LRU cache policy of vLLM's prefix-caching can suffer more from the eviction of the reusable KV context, while the inference could benefit more from \SYS{}. 
For example, when the length of shared prefix is 16000, and the share degree is 16, the token reusing ratio of vLLM and SGLang are 6.3\% and 5.2\%, while the reusing ratio of \SYS{} is 92.6 \%, much better than the other two baselines.

\textbf{\cb{Global Prefix Identification Overhead.}}
We have measured the time of the global prefix identification described in Sec.\ref{subsec:prefix-identification}.
The time range of the overhead starts before adding the token ids of each prompt to form the basic prefix tree and ends after generating the final list of prefix-sharing groups.
As for the industry workload in Sec.\ref{subsubsec:case1}, the total overhead of the 8000 requests is 1.51 seconds.
While the time to process the 8000 requests with the input of prefix-sharing groups format is 919.54 seconds.
The global prefix identification procedure nearly has no overhead compared with the end-to-end execution time.

\subsubsection{\cb{Effect of Request Reordering and Token-Batching}}
\label{subsubsec:breakdown-reordering}

\begin{figure}[]
\centering
\includegraphics[width=1.0\columnwidth]{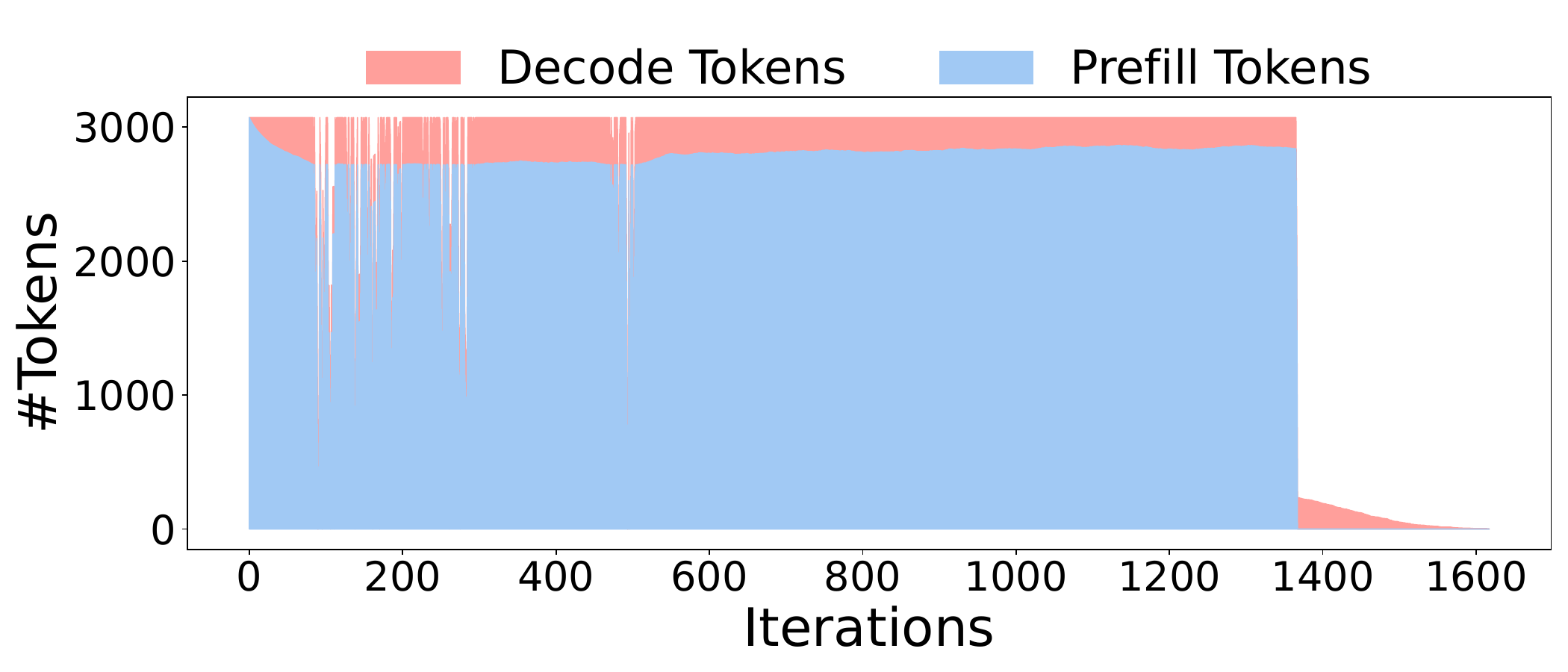}
\caption{Token number per iteration with token-batching optimization (Sec.\ref{subsec:token-batching}), showing significant valley reduction compared to Fig.\ref{fig:valley-example}'s baseline. Prefix-sharing disabled to demonstrate general batched/offline inference efficacy.}
\label{fig:token-batching-token}
\end{figure}

Fig.\ref{fig:token-batching-token} shows the number of per-iteration tokens, using the same workload with Fig.\ref{fig:valley-example}.
It clearly shows that the request reordering and token-batching optimization in \SYS{} successfully reduces the "valleys" of the iterations, thus increasing the overall GPU utilization.

\begin{table}[]
\center
\scriptsize
\caption{Token-Batching Breakdown.}
\begin{tabular}{@{}l|c|cccc@{}}
\toprule
Settings                    & Disable token-batching & Enable token-batching  \\ \midrule
vLLM + c                    & 5.79              &        \textbf{5.94}          \\ \midrule
\SYS{}            & 8.16              &   \textbf{8.47}          \\ \bottomrule   
\end{tabular}
\label{table:token-batching-anal}
\end{table}

We evaluated our token-batching optimization through ablation studies on throughput-oriented token-batching. Testing on Qwen-2.5-7B with 3000 industry workload query (Fig.\ref{table:token-batching-anal}) shows effective performance improvement. Results demonstrate effectiveness both with and without prefix-sharing, with combined optimizations working particularly well in prefix-sharing scenarios.

\subsubsection{\cb{Effect of Horizontal Kernel Fusion}}
\label{subsubsec:breakdown-kernel-fusion}

The existing works use separate kernels to compute the partial Attention on the prefix and the distinct KV context.
This will lead to the tail effect of each kernel and the launch overhead of these kernels.
\SYS{} horizontally fuses the two partial Attention calculation into one kernel.
Different parts of the computation are performed in different thread blocks.
Note that the problem shape of the two parts are different.
\SYS{} applies different tiling configuration for the two parts to achieve the best performance.
It uses the common auto-tuning method to find the best configuration.

We implement the Attention in \SYS{} through OpenAI Triton language~\cite{triton}.
We use the builtin autotuner of Triton to tune the tiling size of the Attention implementation in \SYS{}.
Besides that, some ahead-of-time compilation methods are applied to reduce the launching overhead, including warm-up process on NVIDIA GPUs, and AOTriton\footnote{https://github.com/ROCm/aotriton} on AMD GPUs.

\begin{figure}[!ht]
    \centering

	\subfloat[ Kernel performance under \textit{decoding} scenarios, \#request = 32.]{
		\begin{minipage}[b]{1\columnwidth}
        \includegraphics[width=1\textwidth]{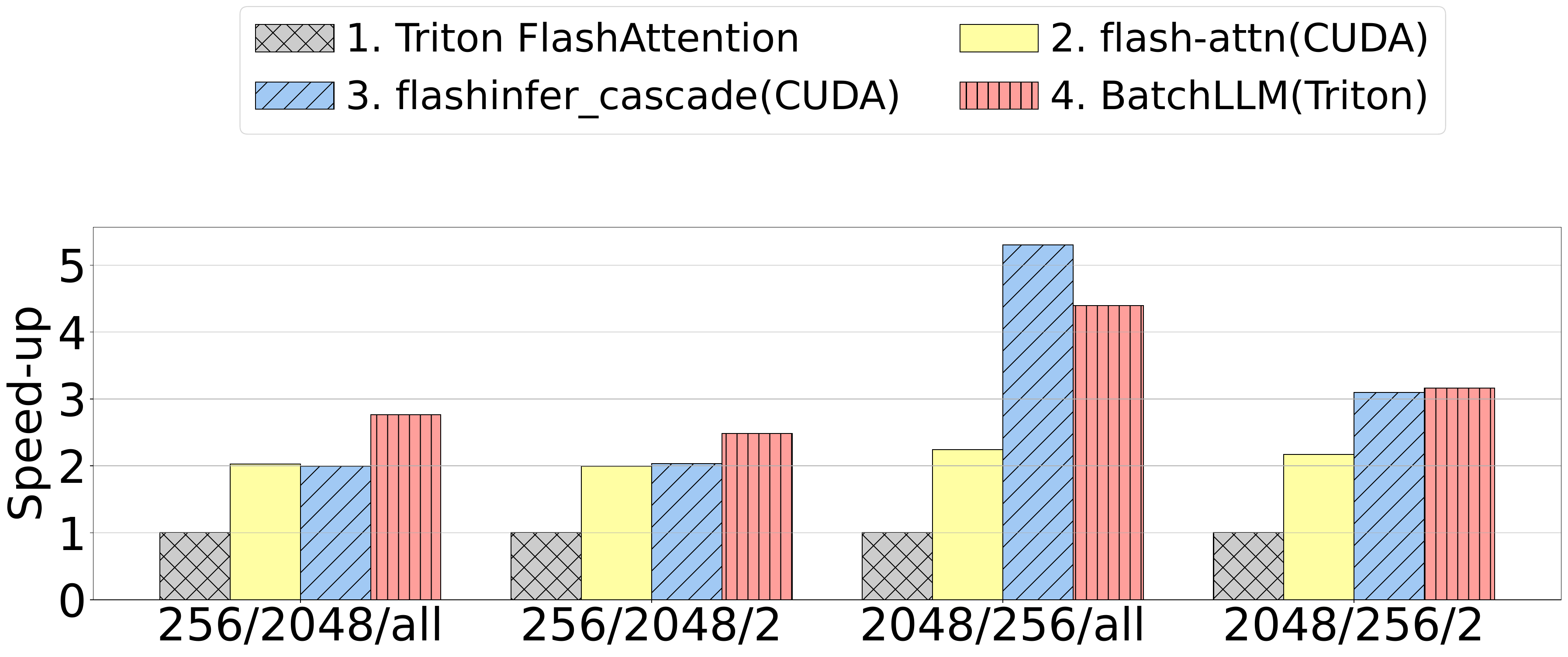}
		\end{minipage}
	    \label{fig:kernel-decoding_32}
	} \\

    \subfloat[ Kernel performance under \textit{decoding} scenarios, \#request = 256.]{
		\begin{minipage}[b]{1\columnwidth}
  	 	\includegraphics[width=1\textwidth]{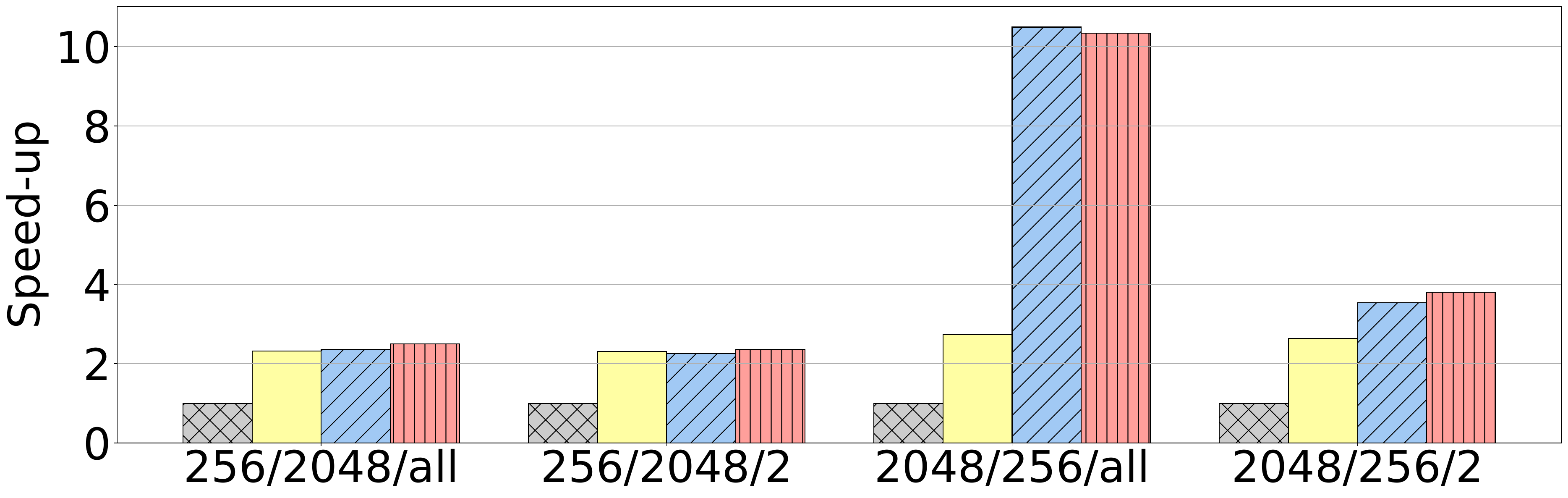}
		\end{minipage}
	    \label{fig:kernel-decoding_256}
	} \\

     \subfloat[ Kernel performance under \textit{chunked-prefill} scenarios, with \#prefill-request = 7, and \#decoding-request = 256.  ]{
		\begin{minipage}[b]{1\columnwidth}
			\includegraphics[width=1\textwidth]{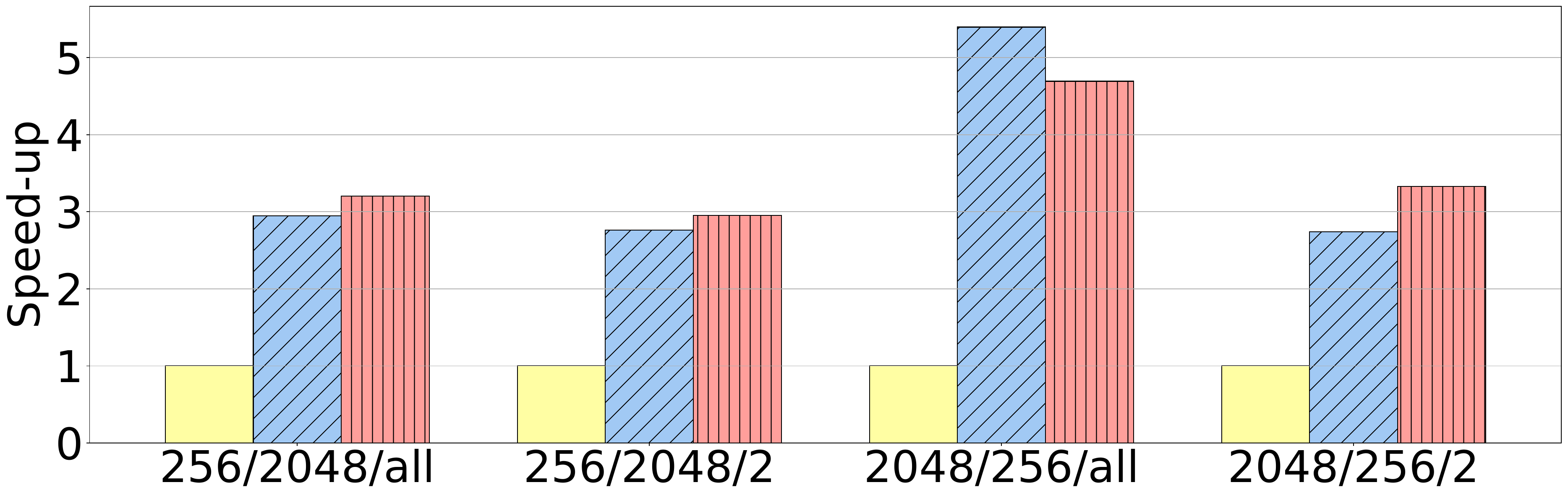}
		\end{minipage}
		\label{fig:kernel-chunk}
	} \\
    \caption{The performance comparison between the baseline, CascadeInfer and \SYS{}.
    The setting \textit{i/j/k} (like \textit{256/2048/all}) means the length of \textit{global shared prefix}, the length of \textit{distinct parts} for each request, and \textit{how many requests} in one \textit{prefix-sharing group}. }
    \label{fig:kernel-performance}
\end{figure}

We compare \SYS{}'s prefix-shared Attention implementation with several baselines on both NVIDIA A100 GPU and AMD MI200 GPU. 
Fig.\ref{fig:kernel-performance} shows the performance comparison of different implementations, 
including Triton official FlashAttention implementation, 
official FlashAttention v2.6.1, 
FlashinferCascade v0.1.4~\cite{cascade-inference}, and \SYS{}'s kernels.
We compare the performance among different kernels under two scenarios: the pure decoding scenario (request count 256) and the chunked-prefill scenario where a token-batch includes both prefill chunks and decoding tokens (7 prefill requests and 256 decoding requests).

Results show that the Triton-based FlashAttention still has a performance gap to the CUDA-based FlashAttention, although we have already tuned the hyparameters using Triton's autotuner.
This indicates the high-level Triton compiler still have a room to achieve the performance of low-level CUDA compiler for complex computations.

However, based on the insufficient performance of Triton, our kernel still shows good performance compared to the Cascade-Inference kernel.
\SYS{}'s kernel can still achieve similar or better performance comparing with all these CUDA implementations,
showing the potential of the kernel optimization in \SYS{}.

\subsection{\cb{Limitations and Deployment Considerations}}

\cb{
While \SYS{} demonstrates superior performance in the industrial scenarios detailed in Sec.\ref{subsec:scenarios}, it is subject to specific limitations.
Primarily, the system's design may adversely affect individual request latency, rendering it less suitable for latency-critical workloads. A potential mitigation is the development of a hybrid architecture that dynamically toggles between streaming and batching modes. Specifically, when the system encounters a backlog of tasks—transitioning into a batching-intensive regime—and identifies a high global prefix sharing ratio, it could employ \SYS{} techniques to maximize throughput and reduce queue depth. Otherwise, the system should utilize standard execution paths to prioritize low latency. We leave the implementation of such a hybrid system for future work.
Secondly, the advantages of \SYS{} are diminished in environments lacking common prefixes. In the absence of prefix caching, performance gains are restricted solely to those provided by token batching. Therefore, we recommend deploying \SYS{} specifically for prefix-heavy, throughput-oriented workloads rather than general-purpose tasks.
}

\section{Related Work}
\label{sec:related}

\textbf{Prefix sharing systems.}
The vLLM~\cite{vllm} proposes the PagedAttention to manage the KV memory in blocks, which enables to reuse the KV context both intra and inter requests in memory block level.
Prompt Cache~\cite{promptcache} and SGLang~\cite{sglang} propose the domain specific language (DSL) to define the shared prefix of prompt.
Besides, SGLang proposes the radix tree based KV cache management with LRU policy for KV  memory eviction.
RAGCache~\cite{ragcache} studies the common KV caching for RAG optimization.
Some recent works~\cite{cachedattention,pdserve,mooncake,memserve,openai-prompt-cache} study the distributed request scheduling with the consideration of prompt prefix reuse.
None of these works identify the common prefix ahead-of-time and enable the explicit reusing.
Instead, \SYS{} manages the prefix reusing explicitly and can reuse the KV context the best for the large batched scenario.

\textbf{Token batching and serving optimization.}
Orca~\cite{orca} proposes the continuous batching method for transformer models with the per-iteration token batching of auto-regressive models~\cite{cellular-batching}.
FastGen~\cite{splitfuse} and SARATHI~\cite{sarathi} study the mix of prefill chunks and decoding tokens into the same token-batch to increase the compute intensity of the batch.
FlexGen~\cite{flexgen} proposes the swizzled token computation and offloading to support the LLM inference with limited GPU memory.
Some works~\cite{learning-to-rank, fairness} study the request scheduling to achieve better SLO for online LLM serving.
\SYS{} differs from these works as it targets the large batched inference, leveraging the static information to reorder the requests and using the memory-aware scheduling to enlarge the size of token-batch.
The vLLM~\cite{vllm} uses the multi-step scheduling method to reduce the token-batch scheduling overhead, which is orthogonal to \SYS{}.
Another orthogonal dimension to improve the token-batch performance is the pruning and quantization~\cite{flash-llm, gptq, awq}, which can be applied concurrently with the optimizations in this paper.

\textbf{Attention kernel optimization.}
Memory-efficient Attention~\cite{memory-efficient-attention} and FlashAttention~\cite{flashattention} fuse the Attention operators into a single kernel to boost the performance, with Online Softmax~\cite{onlinesoftmax} to tile the Attention computation.
The recent prefix-shared Attention optimization (ChunkAttention~\cite{chunkattention}, RelayAttention~\cite{relayattention}, Hydragen~\cite{hydragen}, and Cascade Inference~\cite{cascade-inference}) compute the attention on the prefix and other part separately in different kernels, for which the former is transformed from matrix-vector multiplication between Q and K/V into MatMul, and reduce the results of different parts through Online Softmax.
Different from these works, \SYS{} fuses the Attention computation of different parts into the same kernel to reduce tail latency and kernel launch overhead.
The horizontal fusion has been proposed and used in the existing machine learning optimizers~\cite{recom, recflex, hfuse}.
\SYS{} borrows this idea and applies it in the prefix-shared Attention.

\section{Conclusion}
\label{sec:conclusion}

This paper presents \SYS{}, the holistic optimization techniques for large batched LLM-based information processing and management tasks.
It identifies the limitations of existing methods, and optimizes the tasks with the global information of the large batch.
It explicitly extracts the common prefix globally to avoid prefix's early eviction problem, and simplifies the prefix pattern by enlarging the first-level prefix with a DP algorithm on the tree to reduce the overhead of scheduling and Attention computation.
It schedules the requests at the granularity of prefix-sharing groups, which enables the global prefix sharing the best and shrinks the lifetime of prefix's KV memory.
It proposes the request reordering and memory-centric token batching method to better mix the prefill chunks into the decoding token-batches and thus better saturates the GPU.
Finally, it presents the horizontal fused prefix-shared Attention kernel to reduce the tail effect and kernel launch overhead.
Extensive evaluation shows that \SYS{} outperforms vLLM and SGLang by $1.3\times$ to $10.8\times$ on a set of microbenchmarks and a typical industry workload under different hardware environments.

\bibliography{reference}
\bibliographystyle{mlsys2026}

\end{document}